\documentclass[lettersize,journal]{IEEEtran}
\usepackage{amsmath,amsfonts}
\usepackage{algorithmic}
\usepackage{algorithm}
\usepackage{array}
\usepackage[caption=false,font=normalsize,labelfont=sf,textfont=sf]{subfig}
\usepackage{textcomp}
\usepackage{stfloats}
\usepackage{url}
\usepackage{verbatim}
\usepackage{graphicx}
\usepackage{cite}
\usepackage{subfig}
\usepackage{tabularx}  
\usepackage{booktabs}  
\usepackage{multirow}  
\usepackage{balance}
\begin{document}

\title{Multi-Task Learning Using Uncertainty to Weigh Losses for Heterogeneous Face Attribute Estimation\\}


\author{
\IEEEauthorblockN{Huaqing Yu${^{1,2}}$, Yi He${^1}$\thanks{Yi He is the corresponding author.}, Peng Du${^1}$, Lu Song${^1}$}\\
\IEEEauthorblockA{\textit{${^1}$Tianjin Key Laboratory of Intelligent Unmanned Swarm Technology and System, \\School of Electrical and Information Engineering, Tianjin University, Tianjin 300072, China } \\
\textit{${^2}$Tencent, Shenzhen 518000, China} \\
\{huatsing, heyi, dupeng2022, songlu\}@tju.edu.cn}
}

\markboth{XXX,~Vol.~xx, No.~x, xx~xxxx}%
{Shell \MakeLowercase{\textit{Yuan et al.}}: Multi-Task Learning Using Uncertainty to Weigh Losses for Heterogeneous Face Attribute Estimation}


\maketitle

\begin{abstract}
Face images contain a wide variety of attribute information. In this paper, we propose a generalized framework for joint estimation of ordinal and nominal attributes based on information sharing. We tackle the correlation problem between heterogeneous attributes using hard parameter sharing of shallow features, and trade-off multiple loss functions by considering homoskedastic uncertainty for each attribute estimation task. This leads to optimal estimation of multiple attributes of the face and reduces the training cost of multitask learning. Experimental results on benchmarks with multiple face attributes show that the proposed approach has superior performance compared to state of the art. Finally, we discuss the bias issues arising from the proposed approach in face attribute estimation and validate its feasibility on edge systems.
\end{abstract}


\section{Introduction}
\IEEEPARstart{A}{s} an important cue for social interaction, face images provide a variety of salient information. Over the past 20 years, significant advances have occurred in the field of extracting attribute information from face images to determine subject identity. And it is applied to video security surveillance, personality recommendation, social media \cite{1-7proencca2021uu,yin2020fan}, etc.

However, most previous studies \cite{cao2020rank,icmeage} are limited to estimating a single attribute, or learning separate model for each face attribute. Considering the complexity and computational effort, this is not suitable for realistic applications anymore. 

To address these limitations, new approaches have been developed to explore attribute correlations for joint estimation of multiple face attributes. Even these approaches have some serious limitations. For example, Sun et al.\cite{sun2017demographic} used the same features to estimate all attributes of a face, ignoring the heterogeneity between attributes. Han et al.\cite{huhan2017} abandoned lightweighting  to explore correlations between heterogeneous attributes.

\begin{figure}[h]
\centering
  \includegraphics[width=0.45\textwidth]{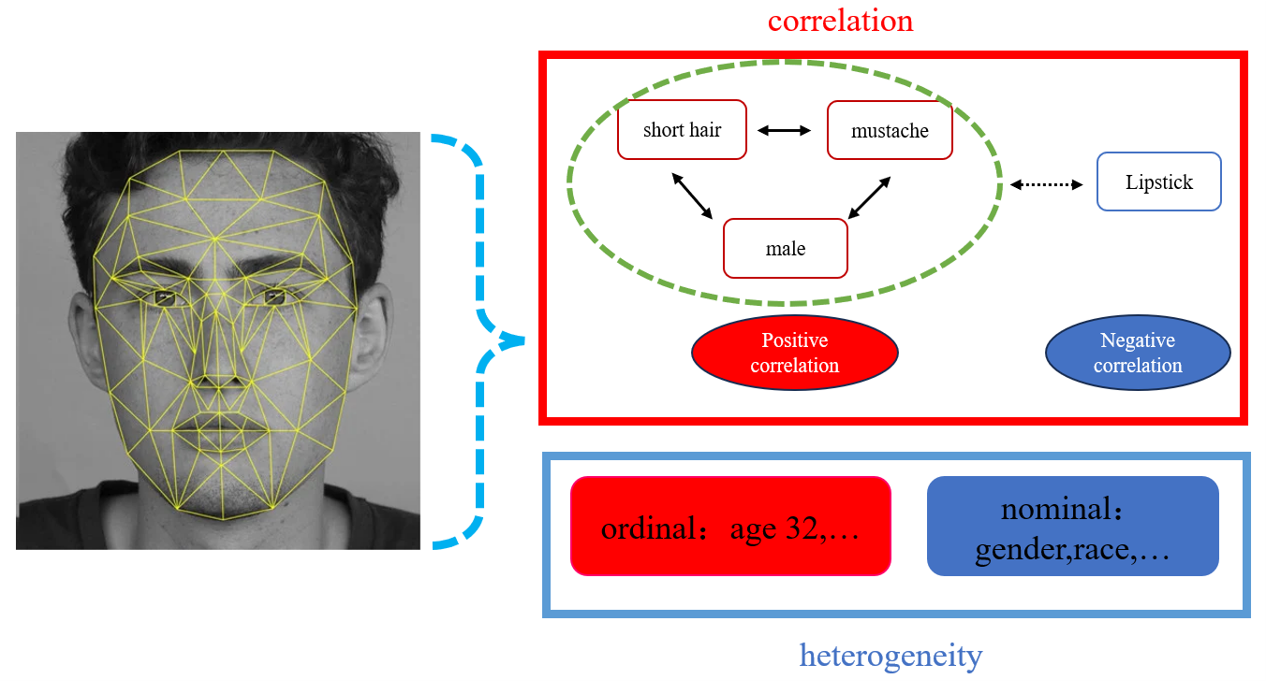}
  \caption{Individual face attributes have both correlation and heterogeneity.}
  \label{face}
\end{figure}

As depicted in Figure~\ref{face}, we observe rich correlations and heterogeneities among multiple attributes of the face. For instance, individuals with a mustache and short hair are more likely to be male, and less likely to wear lipstick. Attributes like age follow an ordered pattern, while gender and race are nominal. While attribute correlation can be utilized to improve the robustness of attribute estimation, attribute heterogeneity should also be tackled by designing appropriate prediction models.

The hierarchical nature that CNNs possess provides a medium for expressing the correlation of heterogeneous face attributes\cite{zeiler2014visualizing}. The lower layers of  CNNs are inclined to capture local common features within the image, such as textures and edges. These features contain relatively less semantic information but exhibit precise target localization. In contrast, the upper layers of CNNs gradually learn category-related features with rich semantic. Therefore,  we propose a multi-task learning approach based on deep CNNs  to take into account the correlation and heterogeneity of face attributes.

In this work, we shared the low-level parameters across all tasks and designed separate classifiers for each task. We focus on ordinal and nominal face attributes, transforming the ordinal regression problem into a linear combination of several binary classification subproblems, which simplifies the task of ordinal attribute estimation.

Specifically, we eschew the conventional approach of manually adjusting multi-task loss weights as observed in previously studies\cite{han2017heterogeneous}. Instead, we construe the homoscedastic uncertainty as a task-dependent weight. This effectively quantifies the noise and scale inherent in each task, facilitating the exploration for optimal weights.

In summary, the key contributions of this paper are:
\begin{itemize}
\item We propose a Deep Multi-Task Learning (DMTL) approach for estimating ordinal and nominal attributes of faces and verify that it is feasible in edge systems;

\item In the estimation of ordinal attributes, we treat the regression problem as a linear combination of a series of binary classification problems, thereby mitigating estimation bias;

\item Based on the homoskedastic uncertainty, we realize the optimal loss-weight search for multiple heterogeneous attribute estimation tasks.
\end{itemize}

\section{Proposed Approach}

\subsection{Network Structure}
\begin{figure*}[h]
\setlength{\fboxsep}{0pt}%
\setlength{\fboxrule}{0pt}%
\begin{center}
\includegraphics[width=5in]{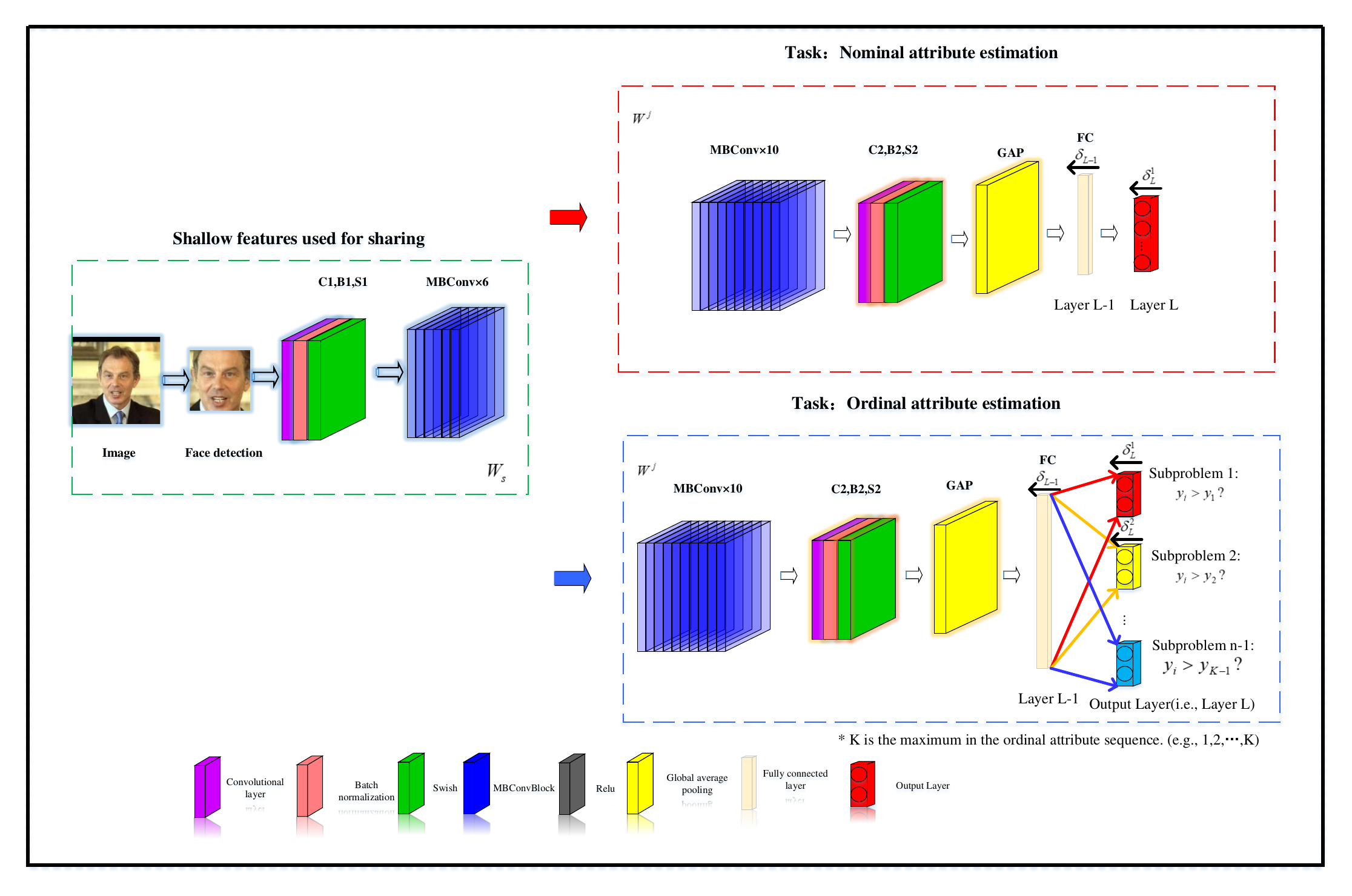}

\end{center}
\caption{Overview of the proposed DMTL approach consisting of an early-stage shared feature learning for all the attributes, followed by category-related feature learning for heterogeneous attribute categories.  We use the MBConv\cite{tan2019efficientnet}  for shared feature learning. Each task achieves optimal estimation of individual heterogeneous attributes through fine-tuning, e.g., nominal versus ordinal.}
\label{F_overview}

\end{figure*}

Multi-task learning is commonly implemented through the sharing of parameters in the hidden layers, either through hard parameter sharing or soft parameter sharing \cite{ruder2017overview}. Baxter\cite{baxter1997bayesian} demonstrated that the risk of overfitting the shared parameters is reduced by an order of magnitude $T$ (where $T$ represents the number of tasks) compared to overfitting the parameters of the output layers. Given that hard parameter sharing enables multiple tasks to share low-level parameters, it possesses advantages such as reduced inference latency and lower storage costs. Therefore, we designed a deep multi-task network based on hard parameter sharing, as illustrated in Figure~\ref{F_overview}.

Suppose a training dataset with $N$ face images, each with $T$  attributes to be  estimated. The dataset is denoted as $D = \left\{ X, Y \right\}$, where $X = \left\{X_i \right\}_{i=1}^N$, and $Y=\left\{\left\{y_i^j \right\}_{j=1}^T \right\}_{i=1}^N$. Minimizing the regularized error function leads to the construction of a multi-task learning model, as depicted in Equation~\ref{eq1}.

\begin{equation}\label{eq1}
{\underset{\{W^j\}_{j=1}^T}{\arg \min} \sum\limits_{j=1}^{T}\sum\limits_{i=1}^{N} \mathcal {L}(y_i^j,\mathcal{F}(X_i,W^j)) + \gamma \Phi (W^j)}
\end{equation}
where $\mathcal{F(\cdot,\cdot)}$ is the attribute estimation function of the input ${X_i}$ and weight vector ${W^j}$; $\mathcal{L}(\cdot,\cdot)$ is the prescribed loss function between the estimated values of ${\mathcal{F}}$ and the corresponding ground-truth values ${y_i^j}$; ${\Phi(\cdot)}$ is the regularization term which penalizes the complexity of weights, and ${\gamma}$ is a regularization parameter ${(\gamma > 0 })$.

In the proposed approach, we employed hard parameter sharing for the low-level features of multiple heterogeneous attribute estimation tasks. Consequently, Equation~\ref{eq1} has been reformulated as Equation~\ref{eq2}.

\begin{equation}\label{eq2}
\begin{split}
 \underset{W_s, \{W^j\}_{j=1}^T}{\arg \min} \sum\limits_{j=1}^{T}\sum\limits_{i=1}^{N} &{\beta}^j \mathcal {L}(y_i^j,\mathcal{F}(X_i,W^j \circ W_s)) \\
& + {\gamma}_1 \Phi (W_s) + {\gamma}_2 \Phi (W^j)
\end{split}
\end{equation}
where$\sum\limits_{j=1}^{T}{\beta}^j = 1$ indicates the weight of loss for each task, ${W_s}$ controls feature sharing among the face attributes, and ${W^j}$ controls update shared features of each face attribute. 

\subsection{Heterogeneous Face Attribute Estimation}
\subsubsection{Nominal Attribute Estimation}

Face nominal attributes are often regarded as a classification problem because they have no intrinsic order among the categories.We choose to use the cross-entropy loss $\mathcal {L}$ to deal with it.

\begin{equation}\label{eq4}
{\mathcal {L} = - \frac{1}{N}\sum\limits_{i=1}^{N}\sum\limits_{c=1}^{C}1\{y_i = \hat{y}_i^c\} \log p(\hat{y}_i^c)}
\end{equation}
\begin{equation}\label{eq5}
\begin{aligned}
p(\hat{y}_i^c) &= \text{Softmax}(\hat{y}_i^c)\\
    &=\frac{e^{{y}_i^c}}{\sum_{c=1}^{C}e^{{y}_i^c}} 
\end{aligned}
\end{equation} 
where $N$ is the total number of samples in the dataset, $C$ is the number of classes, $y_i$ is the  the label for the $i$-th  sample, $\hat{y}_i^c$ is the predicted probability that the $i$-th sample belongs to class $c$. $1\{\cdot\}$ is an indicator function, it equals 1 when $\{\cdot\}$ is true, and it equals 0 otherwise. $p(\hat{y}_i^c)$ is the model's predicted probability that sample $i$ belongs to class $c$.

\subsubsection{Ordinal Attribute Estimation}
Ordinal attributes have a well-defined sequential relationship between different values. In the age estimation, we observe that age is not only ordered but also discrete. In this paper, we transform the ordinal regression problem into a linear combination of binary classification subproblems. This allows for the direct utilization of well-studied classification algorithms.

Specifically, for the $k$-th binary classification subproblem, we construct  training data as  $D^k = {\{x_i, y_i^k\}}_{i=1}^N$, where the $y_i^k \in \{0,1\}$ is a binary class label indicating whether the rank of the $i$-th sample $y_i$ is larger than $r_k$ as follows,
\begin{equation}\label{eq6}
{y_i^k =  
\begin{cases}
1,& \text{ $ if(y_i > r_k) $ } \\
0,& \text{ $ otherwise $ }
\end{cases}}
\end{equation}
where $r_k \in \{r_1,r_2,\cdots,r_{K-1}\}$ is the label of the $k$-th class of subproblem (e.g., in age estimation,$r_1$ denotes age 1, $r_2$ denotes age 2, $\cdots$).

Then the predicted value of the sample $x_i$ is :
\begin{equation}\label{eq7}
\begin{aligned}
&\hat{y}_i^k =  r_p,\\
&p = 1 + \sum\limits_{k=1}^{K-1}{f_k(x_i)}
\end{aligned}
\end{equation}

where $f_k(x_i) \in \{0,1\}$ is the classification result of the $k$-th binary classifier for the sample $x_i$ (i.e., the $k$-th output of ordinal attribute estimation task), $K$ represents the maximum in the ordinal sequence.

Therefore, each binary classification subproblem employs cross-entropy loss. The total loss $\mathcal {L}$ of ordinal attribute estimation task for $N$ input images is expressed as follows,
\begin{equation}\label{eq8}
\begin{aligned}
\mathcal{L} = & -\frac{1}{N}\sum_{i=1}^{N}\sum_{k=1}^{K-1} [y_i^k \cdot \log p(\hat{y}_i^k = 1) \\
& + \left( 1- y_i^k \right) \cdot \log (1 - p(\hat{y}_i^k = 1))]
\end{aligned}
\end{equation}
where $y_i^k \in \{0,1\}$ is the binary label of the $k$-th subproblem,  $p(\hat{y}_i^k = 1)$ represents the probability of predicting as true. 

\subsection{Multi-Task Learning Using Uncertainty to Weigh Losses}

In Bayesian modeling of deep learning, there is inherent aleatoric uncertainty, where homoscedastic uncertainty remains constant across all input data and varies between different tasks\cite{kendall2018multi}.

We assuming $f^W(x)$ represents the output of the network, the classification can be illustrated as follows,
\begin{equation}\label{e13}
\begin{aligned}
p({y}\mid{{{f^W}(x)}}) & = \text{Softmax}({f^W}(x)) \\
& = \frac{\exp({f^W}(x))}{\sum\limits_{{c^\cdot}} {\exp ({f_{c^\cdot}^W(x))}}}
\end{aligned}
\end{equation} 
where ${{f_{c}}^W(x)}$ the $c$-th output element of the ${f^W}(x)$, $W$ is the weights of the neural network, $y$ denotes the output of the network, and $x$ denotes the input to the network.

We introduce a positive scalar $\sigma$ to learn the homoscedastic uncertainty in multi-task:
\begin{equation}\label{e15}
p({y}\mid{{{f^W}(x)}},\sigma ) = \text{Softmax}(\frac{1}{{{\sigma ^2}}}{f^W}(x))
\end{equation} 

The likelihood for this output can then be written as

\begin{equation}\label{e16}
\log p({y}\mid{{{f^W}(x)}},\sigma ) = \log \text{Softmax}(\frac{1}{{{\sigma ^2}}}{f^W}(x))
\end{equation} 
and 

\begin{equation}\label{e17}
\begin{aligned}
\log p(y = c\mid{f^W}(x),\sigma ) &= \frac{1}{{{\sigma ^2}}}{f_c}^W(x)  \\
    & - \log \sum\limits_{{c^\cdot}} {\exp (\frac{1}{{{\sigma ^2}}}{f_{c^\cdot}}^W(x))}
\end{aligned}
\end{equation} 

In the proposed approach, $T$ tasks share a consistent dataset, ensuring that different tasks satisfy the assumption of being independently and identically distributed, as follows,

\begin{equation}\label{e14}
p({y_1}, \ldots ,{y_T}\mid{f^W}(x)) = p({y_1}\mid{f^W}(x)) \ldots p({y_T}\mid{f^W}(x))
\end{equation} 
with model outputs $y_1, \ldots , y_T$, $T$ denotes the number of tasks or the number of attributes.

The joint loss   $\mathcal {L} (W,\sigma_1,\ldots,\sigma_T)$  for multiple tasks of a single sample is given by the following,

\begin{equation}\label{e18}
\begin{aligned}
\mathcal {L}(W,&\sigma_1,\ldots,\sigma_T)\\
&= -\log p((y_1,\ldots, y_T \mid f^W(x)), \sigma) \\
&= - \log \prod_{t}^{T} p(y_t \mid f^W(x), \sigma_t) \\
&= - \log \prod_t^T \left( \text{Softmax} (y_t \mid f^W(x), \sigma_t) \right) \\
&= \sum_{t}^{T} \left( \frac{1}{\sigma_t^2} \mathcal {L}_t(W) + \log \frac{\sum_{c^{\cdot}} \exp \left( \frac{1}{\sigma_t^2} f_{c^{\cdot}}^W(x) \right)}{\left( \sum_{c^{\cdot}} \exp (f_{c^{\cdot}}^W(x)) \right)^{\frac{1}{\sigma_t^2}}} \right)
\end{aligned}
\end{equation}
where
\begin{equation}\label{e19}
\begin{aligned}
{\mathcal {L}_t}(W) &= -\log  \frac{\exp({f^W}(x))}{\sum\limits_{{c^\cdot}} {\exp ({f_{c^\cdot}^W(x))}}}\\
&= -\log \text{Softmax} ({y_t};{f^W}(x))
\end{aligned}
\end{equation}

For simplification, we rewrite Equation~\ref{e18} as follows,

\begin{equation}\label{e20}
\mathcal {L}(W,{\sigma _1},...,{\sigma _T}) = \sum\limits_t^T {(\frac{1}{{\sigma _t^2}}{\mathcal {L}_t}(W) + \log } {\sigma _t})
\end{equation}
when $\sigma_t \to 1$.

For a batch of samples, $\mathcal {L}_t$ is computed by Equation~\ref{eq4} when $t$-th task is nominal attribute estimation task, $\mathcal {L}_t$ is computed by Equation~\ref{eq8} when $t$-th task is ordinal attribute estimation task.


\subsection{Implementation Details}
For all the face images, we normalize the face images into $224 \times 224 \times 3 $ (height $\times$ width $\times$ channels) based on five facial landmarks. 


We constructed the proposed DMTL network by PyTorch and trained it on Nvidia Tesla K40c by transfer learning. We perform multi-task training using Algorithm \ref{alg:alg}. For the optimization of the loss function, we adopt batch stochastic gradient descent algorithm. The total number of training epochs was set to 80, the learning rate was set to 0.001, the weight decay value was set to 0.0005, batch size to 32, and the momentum coefficient was set to 0.9. 

\begin{algorithm}[!h]
    \caption{Update weights and $\sigma$ of tasks}
    \label{alg:alg}
    \renewcommand{\algorithmicrequire}{\textbf{Input:}}
    \renewcommand{\algorithmicensure}{\textbf{Output:}}
    \begin{algorithmic}[1]
        \REQUIRE training data $D = \{{x_i},y_{i1},\ldots, y_{iT}\}_{i=1}^N$ 
        \ENSURE $\{ {\hat y_{j1},\ldots, \hat y_{jT}}\}_{j = 1}^{N'}$    

        \FOR{$e \in (1,epoch) $}
            \FOR{$t \in (1,T) $}
            \STATE  $\mathcal{L}_t$ $\leftarrow$ Equation~\ref{eq4} if  nominal task else Equation~\ref{eq8}
            \ENDFOR
            \STATE Joint Loss $\mathcal{L}$ using Equation~\ref{e20}
       \STATE update $W$, $\log \sigma^2$

          
        \ENDFOR
        
    \end{algorithmic}
\end{algorithm}

In practice, we train the proposed approach to predict the log variance $\log \sigma^2$ in Equation~\ref{e20}. This avoids any division by zero and more numerically stable.

\section{Experiments And Discussion}

\subsection{Comparison With other Approaches}

\subsubsection{Comparison of performance on Adience}
The Adience\cite{1-1eidinger2014age} benchmark includes age and gender attribute labels. Table~\ref{Tadience_com} presents the accuracy of different approaches. The results show that the proposed approach outperforms the existing approaches in terms of accuracy on two different attributes, reaching 86.59\% and 95.75\%, respectively.

The proposed DMTL approach has better generalization performance compared to single-task learning, and the proposed approach also has performance advantages when compared to other state-of-the-art approaches. For example, GRA-Net treats the age task as a combination of multi-classification and regression problems, while the proposed approach constructs a series of binary classification subproblems to solve the real age through multi-output CNNs, which puts more emphasis on information sharing and utilization.

\subsubsection{Comparison of performance on UTKFace}
The UTKFace\cite{utkface} benchmark includes three facial attributes: age, gender, and race, where the age attribute is an exact integer value. The accuracy of each attribute recognition on UTKFace is listed in Table~\ref{Tutkfaceacc}. The results demonstrate the superiority and feasibility of the proposed approach. 

\begin{table}[h]
\centering
\caption{Comparison of the accuracy of different approaches on the Adience benchmark (in \%)}

    \begin{tabular}{ccc}
    \toprule[1.5pt]
    Approach& Age & Gender\\
    \midrule[1pt]
    
    Single Task & {73.25} & 90.67 \\
    GRA-Net\cite{GraNet} & {66.10} & 81.80 \\
    AFA-Net\cite{AFAnet} & {80.40} & 90.00 \\
    Zhang et al., \cite{KEZHANG} & {70.90} & 90.03\\
    Eidinger et al.,\cite{eidinger2014age} & \text{45.10} & {76.10}\\
    Earnest Paul et al.,\cite{IjjinaAge} & {70.80} & {82.92}\\
    Insha Rafique et al.,\cite{RafiqueAge} & {79.30} & {50.70}\\
    Anweasha Saha et al.,\cite{saha2023age} & {75.10} & {84.94}\\
    
    \textbf{Proposed} & \textbf{86.59} & \textbf{95.75} \\

    \bottomrule

    \end{tabular}\\[10pt] 

\label{Tadience_com}
\end{table}

\begin{table}[h]
\centering
\caption{Comparison of the accuracy of different approaches on the UTKFace benchmark (in \%)}

    \begin{tabular}{cccc}
    \toprule[1.5pt]
    Approach & Age$^1$ & Gender & Race\\
    
    \midrule[1pt]
    Single Task & 60.17 & 90.09 & 76.23\\
    ResNet\cite{Resnet} & 58.6 & 87.4 & -\\ 
    AlexNet\cite{alexNet} & {55.96} & 85.1  & -\\
    DenseNet\cite{DenseNet} & {59.22} & 87.28  & -\\
    VGGNet\cite{VGG} & {57.06} & 83.63  & -\\
    
    Jamoliddin et al., \cite{jamoliddin2022age} & {61.34} & 88.67  & -\\
    \textbf{Proposed} & \textbf{64.74} & \textbf{90.91} & \textbf{79.98} \\

    \bottomrule
    \multicolumn{4}{p{3in}}{\raggedright\small $^1$results are reported using the accuracy with age group. \\‘-’ denotes unavailable metrics in related literature.} \\
    \end{tabular}\\[10pt] 
\label{Tutkfaceacc}
\end{table}

\subsubsection{Comparison of estimation error  in  ordinal attribute tasks}
\begin{table}[h]
\centering
\caption{Comparing the estimation error of different approaches in age estimation task}

\begin{tabular}{ccccc}
\toprule[1.5pt]
\multirow{2}{*}{Approach} & \multicolumn{2}{c}{Adience} & \multicolumn{2}{c}{UTKFace} \\
& MSE & MAE & MSE & MAE \\
\midrule[1pt]
Single Task & 0.9266 & 0.4194 & 116.87 & 6.73\\
Li et al., \cite{li2021learning} & - & 0.41 & - & -\\
MobileNet v2\cite{sandler2018mobilenetv2} & - & 0.37 & - & 7.29\\
Sumit Kothari\cite{Kothariutk} & - & - & - & 6.76\\
FaceNet\cite{FaceNet} & - & - & - & 6.71\\
\textbf{Proposed} & \textbf{0.2409} & \textbf{0.1639} & \textbf{66.3958} & \textbf{5.323} \\
\bottomrule
\multicolumn{5}{p{3in}}{\raggedright\small ‘-’ denotes unavailable metrics in related literature.} \\

\end{tabular}
\label{Tageae}

\end{table}

For ordinal attribute estimation task, we computed the Mean Square Error (MSE) and Mean Absolute Error (MAE) for the age estimation task in Table~\ref{Tageae}. The proposed approach exhibits lower estimation errors compared to other approaches due to the improved generalization performance induced by DMTL.

\subsection{Ablation Study}

\begin{figure}[h]
\centering
  \subfloat[]{\includegraphics[width=0.45\textwidth]{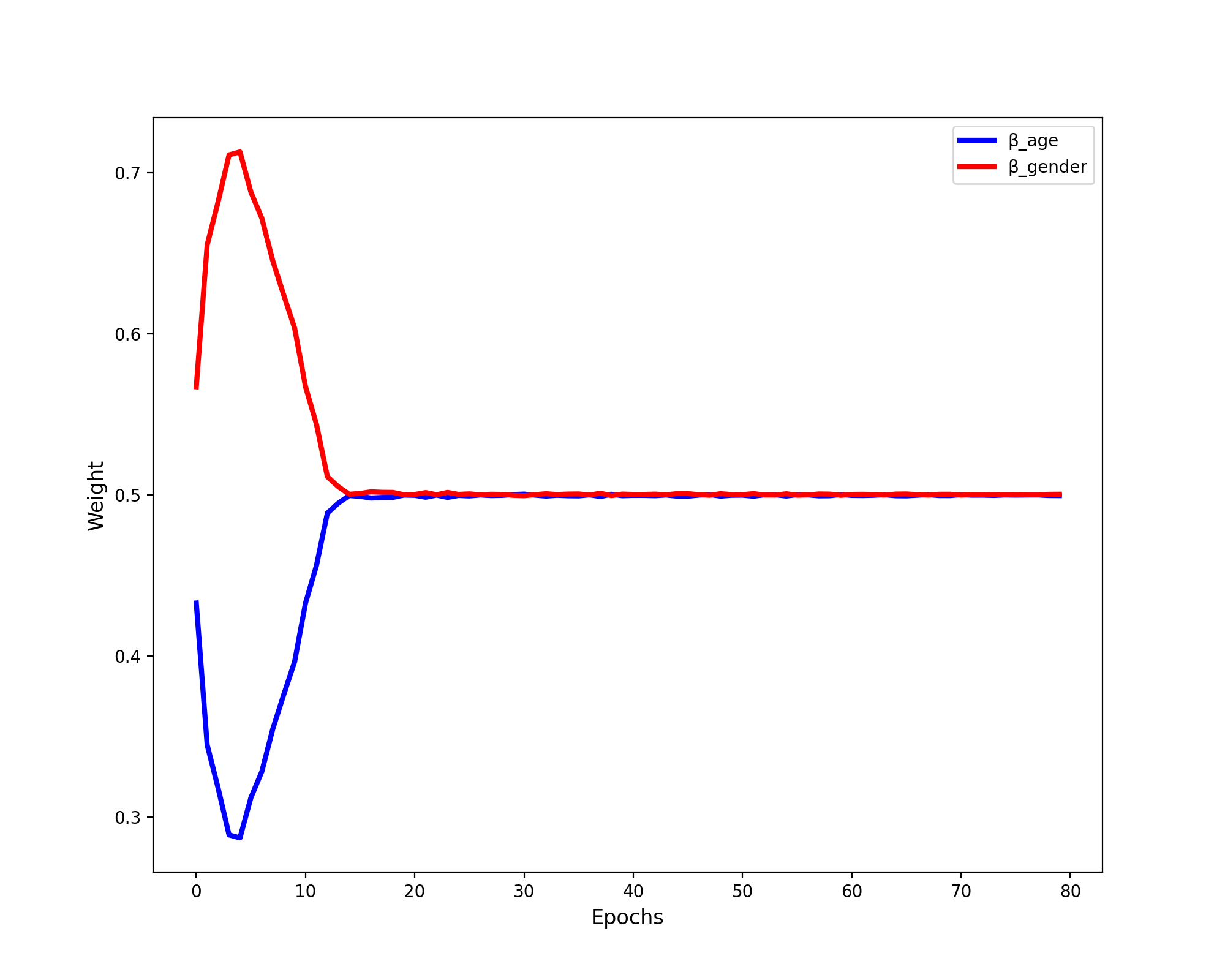}}
  \
  \hfil
  \subfloat[]{\includegraphics[width=0.45\textwidth]{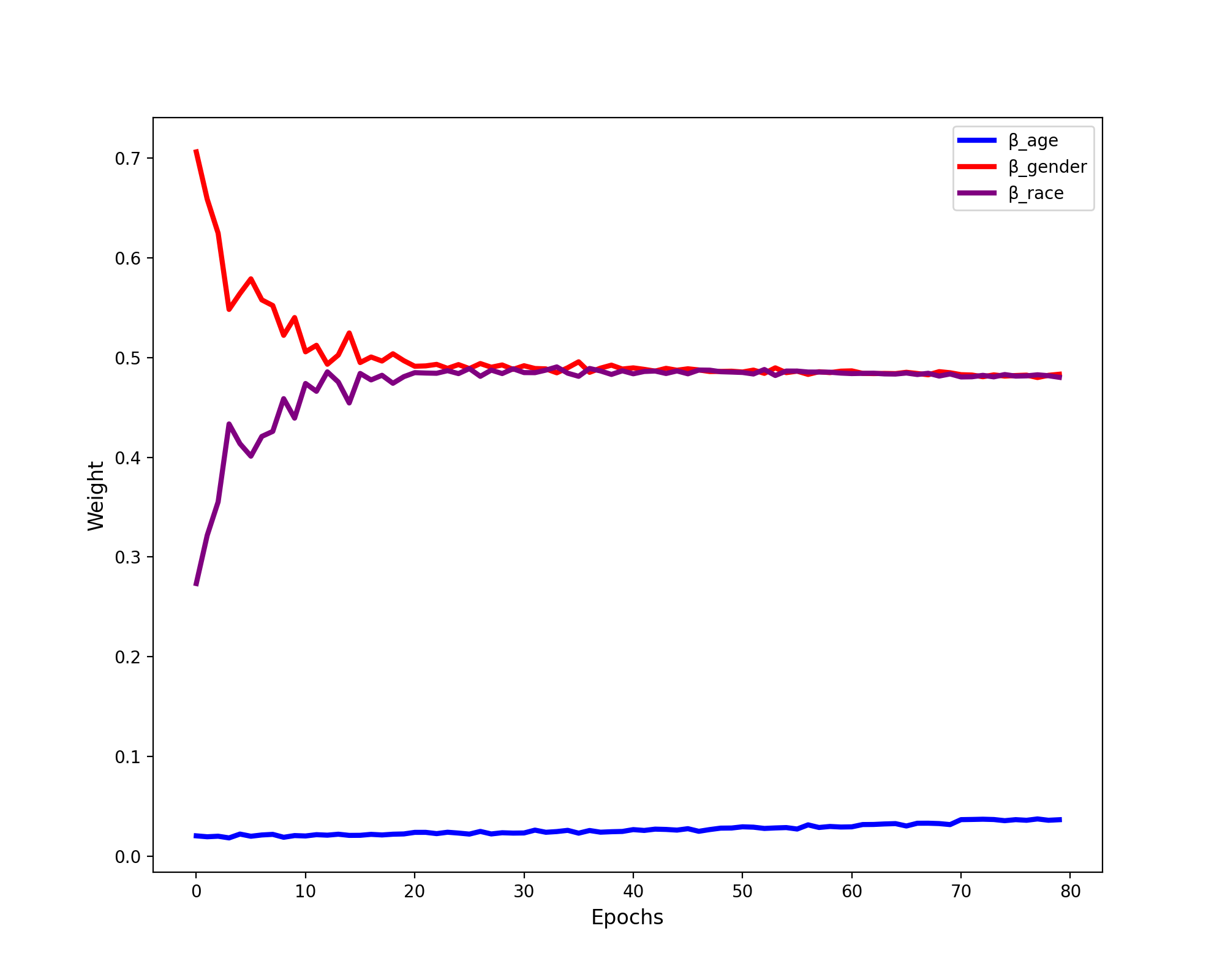}}

  \caption{Convergence curves of task weights training on two datasets, $\beta_i$  represents the weight of the attribute $i$ estimation task in the Joint loss $\mathcal{L}$. (a) Adience. (b) UTKFace. }
  \label{lossweight}
\end{figure}

As shown in Figure~\ref{lossweight}, we depict the variation in weights of task losses during the training process using homoscedastic uncertainty, where $\beta_i$ is normalized using  Equation~\ref{e312}.

\begin{equation}\label{e312}
{
\begin{cases}
&\beta_i = \frac{1}{\sum_{j=1}^M \frac{\sigma_i^2}{\sigma_j^2}}\\
&\sum_{i=1}^M  \beta_i = 1
\end{cases}
}
\end{equation}
where $M$ represents the number of tasks that need to be estimated.

It can be observed that on the Adience dataset, the weight for the age estimation task $\beta_{age}$ is comparable to that of the gender recognition task $\beta_{gender}$, while on the UTKFace dataset, the weight for the age estimation task $\beta_{age}$ is very low. To ensure that tasks with different magnitudes of loss values are adequately trained, Equation~\ref{eq2} or Equation~\ref{e20} assigns smaller weights to tasks with larger losses.

We present the results of ablation experiments in Table~\ref{Tablation}. In the ``Proposed without ordinal opti'' approach, we neglected the special treatment for ordinal tasks, while in the ``Proposed without uncertainty'' approach, we selected the approximate optimal weights for multi-task learning through manual adjustments.

\begin{figure}[h]
\centering
  \includegraphics[width=0.45\textwidth]{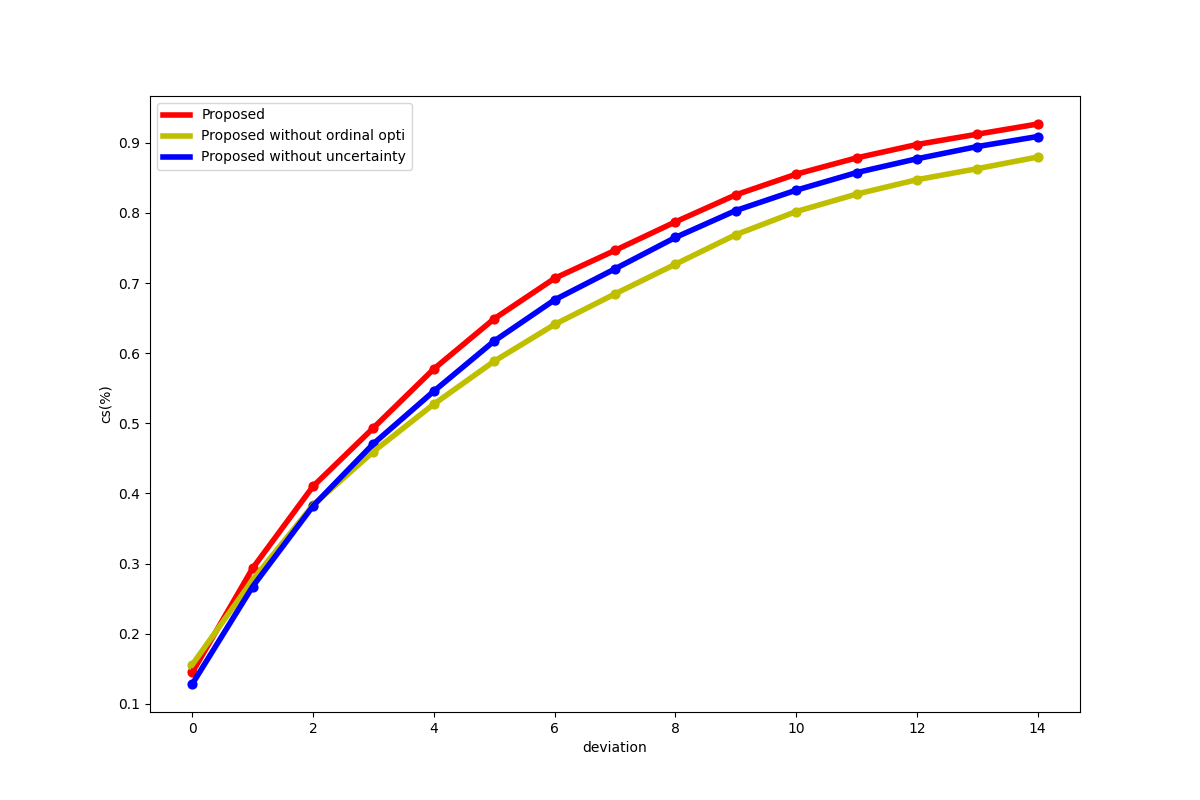}
  \caption{The comparison of age estimation with CS metric on UTKFace benchmark.}
  \label{3-cs}
\end{figure}

For ordinal task, we computed the Cumulative Score (CS) for three approaches on the UTKFace benchmark and illustrated it in Figure~\ref{3-cs}. This will help us to validate different approaches in ordinal attribute prediction performance.

The CS is calculated as follows:

\begin{equation}\label{e21}
CS(i) = \frac{{{N_i}}}{N} \times 100\%
\end{equation}
where $N$ is the total number of test images, and $N_i$  is the number of test images whose the absolute error between the estimated value and the label is less than order $i$.

\begin{table*}[h]
\centering
\caption{Ablation study on Adience and UTKFace benchmarks}

\begin{tabular}{cccccc}
\toprule[1.5pt]
\multirow{2}{*}{Approach} & \multicolumn{2}{c}{Adience} & \multicolumn{3}{c}{UTKFace} \\
& Age$^1$ & Gender & Age$^1$ & Gender & Race \\
\midrule[1pt]
Proposed without ordinal opti & 82.78$|$0.2397 & 94.08 & 61.78$|$6.6190 & 90.63 & 78.49\\
Proposed without uncertainty  &83.70$|$0.2179 & 94.13 & 62.48$|$5.9562 & 90.70 & 78.80\\
\textbf{Proposed} & \textbf{86.59$|$0.1639} & \textbf{95.75} & \textbf{64.74$|$5.3230} & \textbf{90.91} & \textbf{78.98}\\

\bottomrule
\multicolumn{6}{p{4in}}{\raggedright\small $^1$results are reported in terms of both the accuracy with age group and MAE. } \\

\end{tabular}
\label{Tablation}

\end{table*}

Through ablation experiments, we demonstrate the effectiveness of reducing estimation errors by transforming ordinal task into a linear combination of binary classifications subproblems. Furthermore, adjusting the weights of multi-task loss through uncertainty is beneficial.

\subsection{Case Study}
We attempt to utilize the Gard-CAM\cite{selvaraju2017grad} algorithm to explain why the proposed approach is effective.

\begin{figure}[h]
    \centering
\subfloat[]{\includegraphics[width=0.15\textwidth]{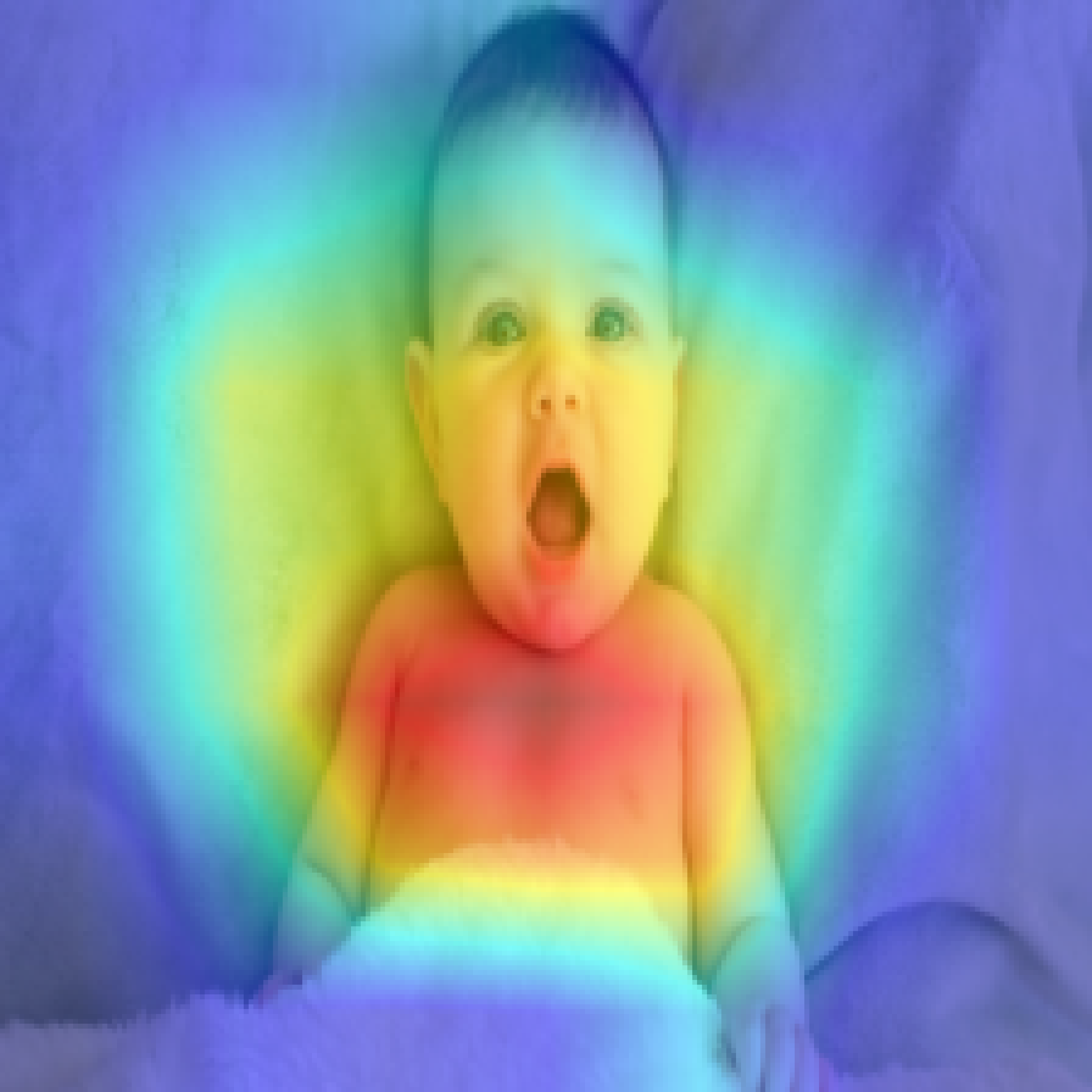}}
  \hfill
  \subfloat[]{\includegraphics[width=0.15\textwidth]{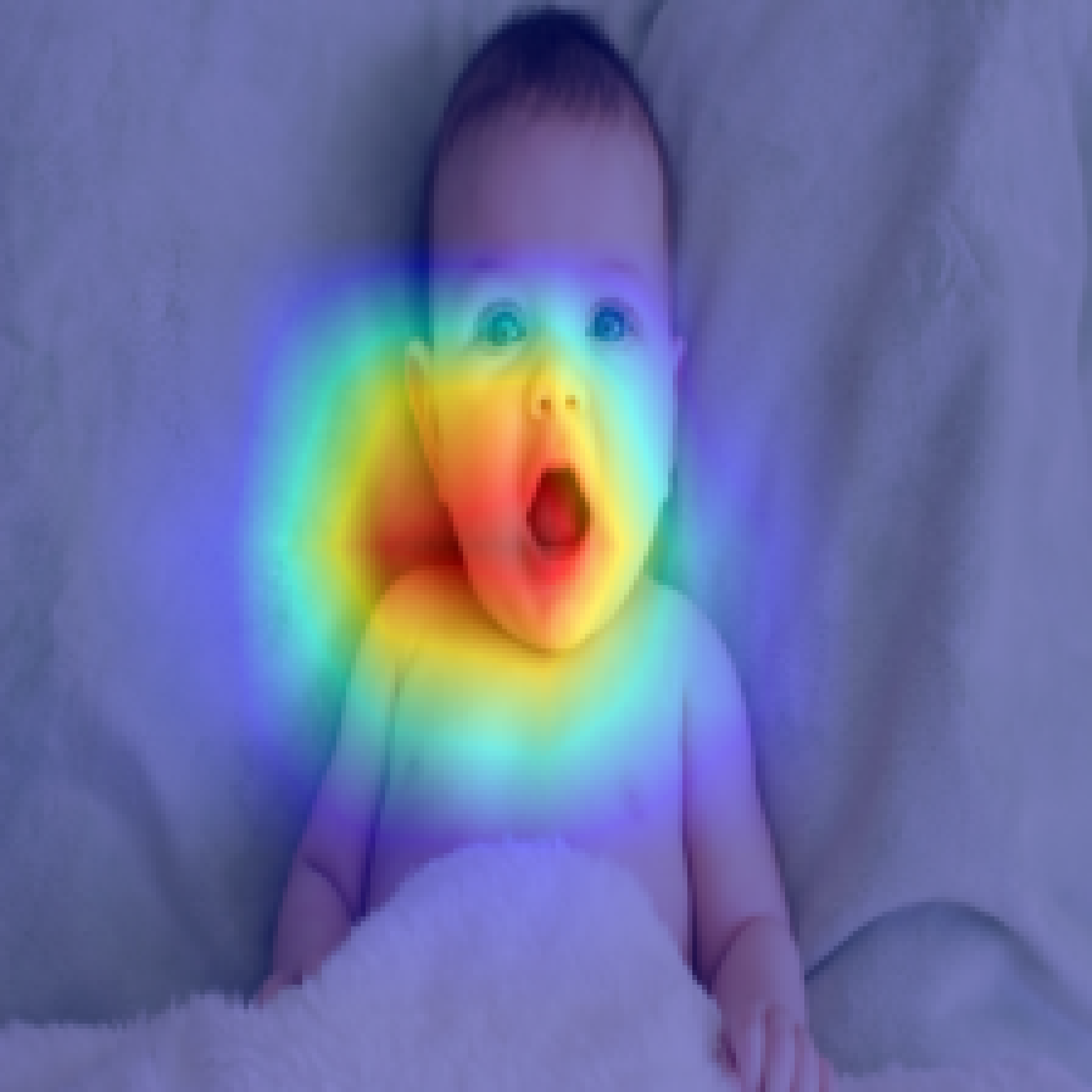}}
  \hfill
  \subfloat[]{\includegraphics[width=0.15\textwidth]{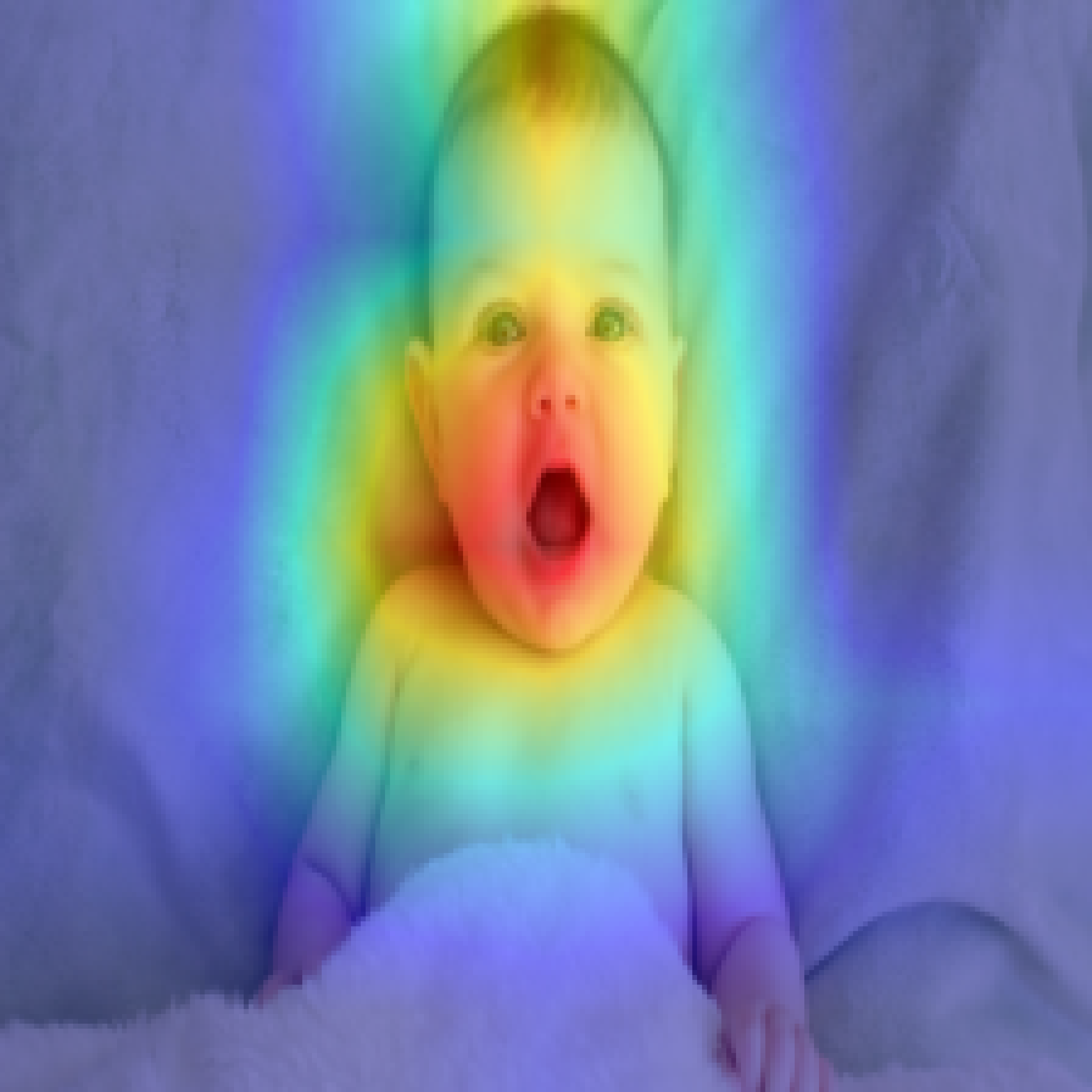}}
  \newline
  \subfloat[]{\includegraphics[width=0.15\textwidth]{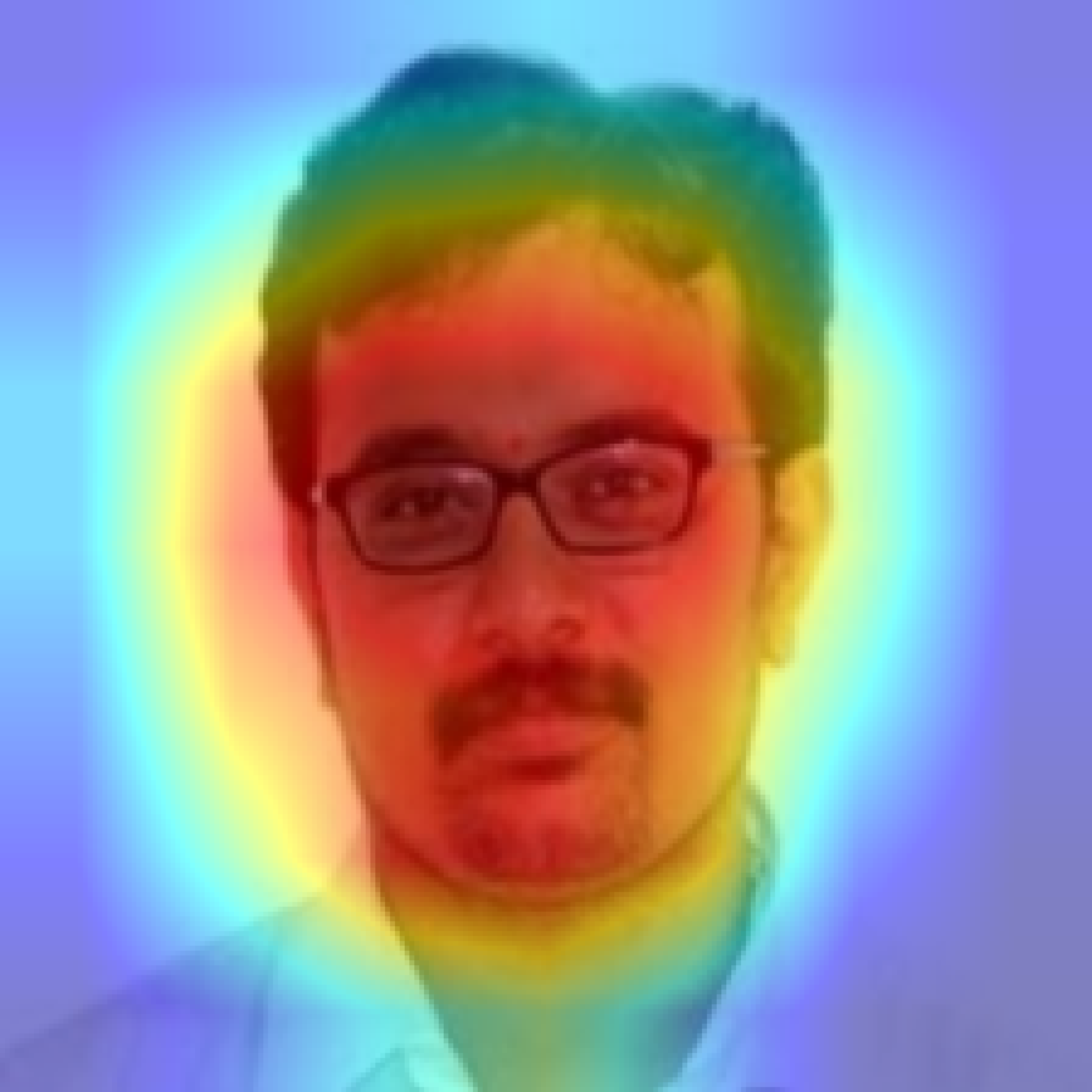}}
  \hfill
  \subfloat[]{\includegraphics[width=0.15\textwidth]{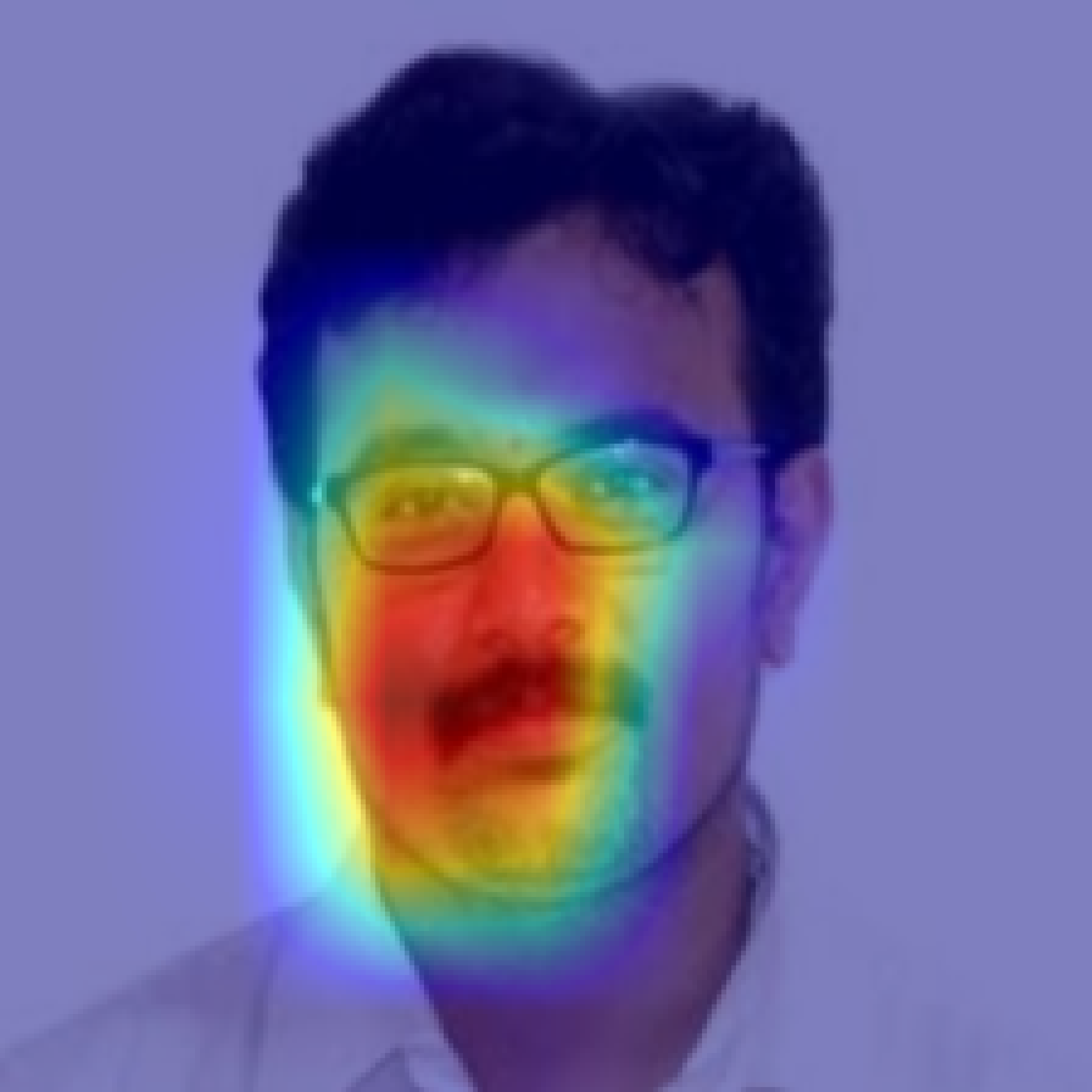}}
  \hfill
  \subfloat[]{\includegraphics[width=0.15\textwidth]{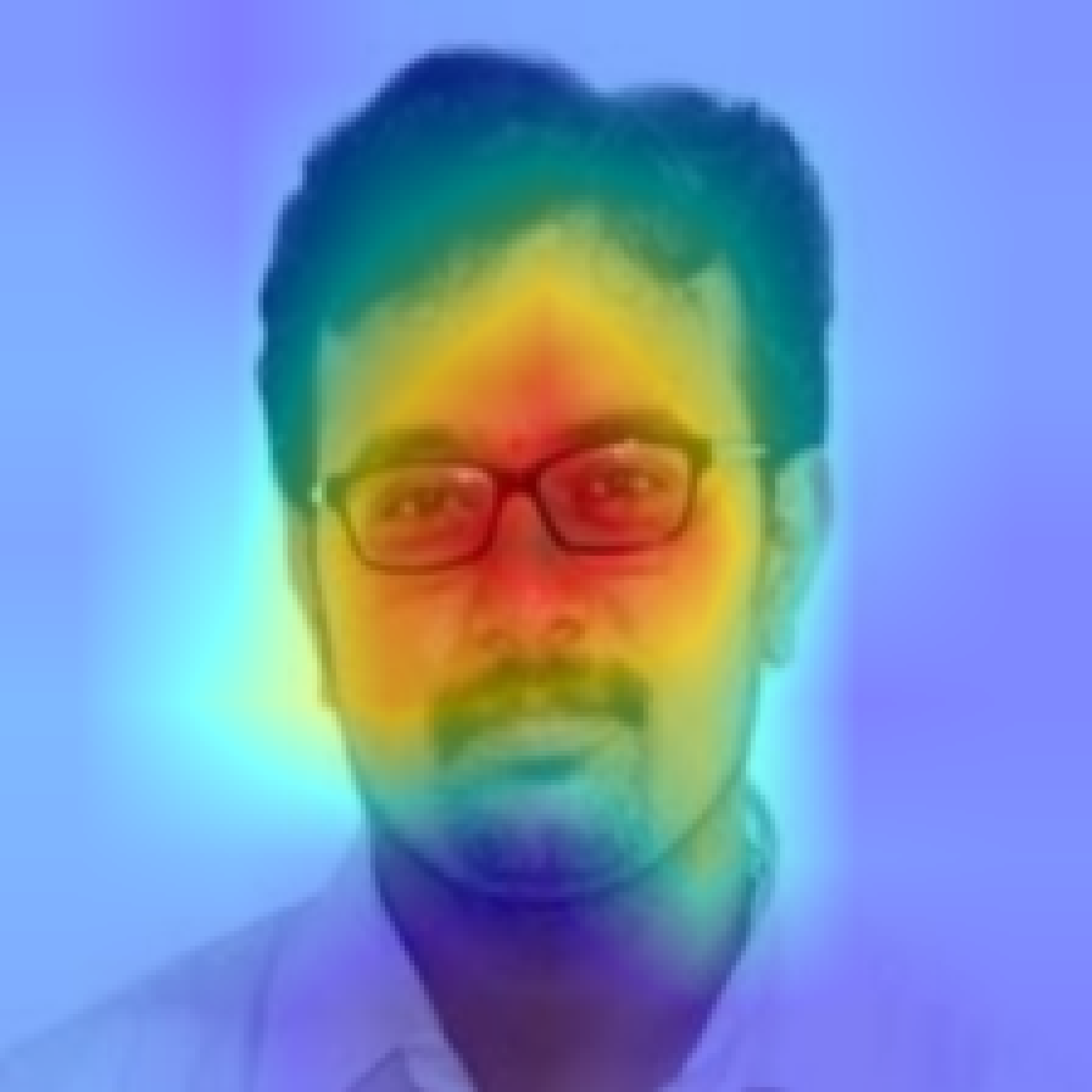}}
  \newline
  \subfloat[]{\includegraphics[width=0.15\textwidth]{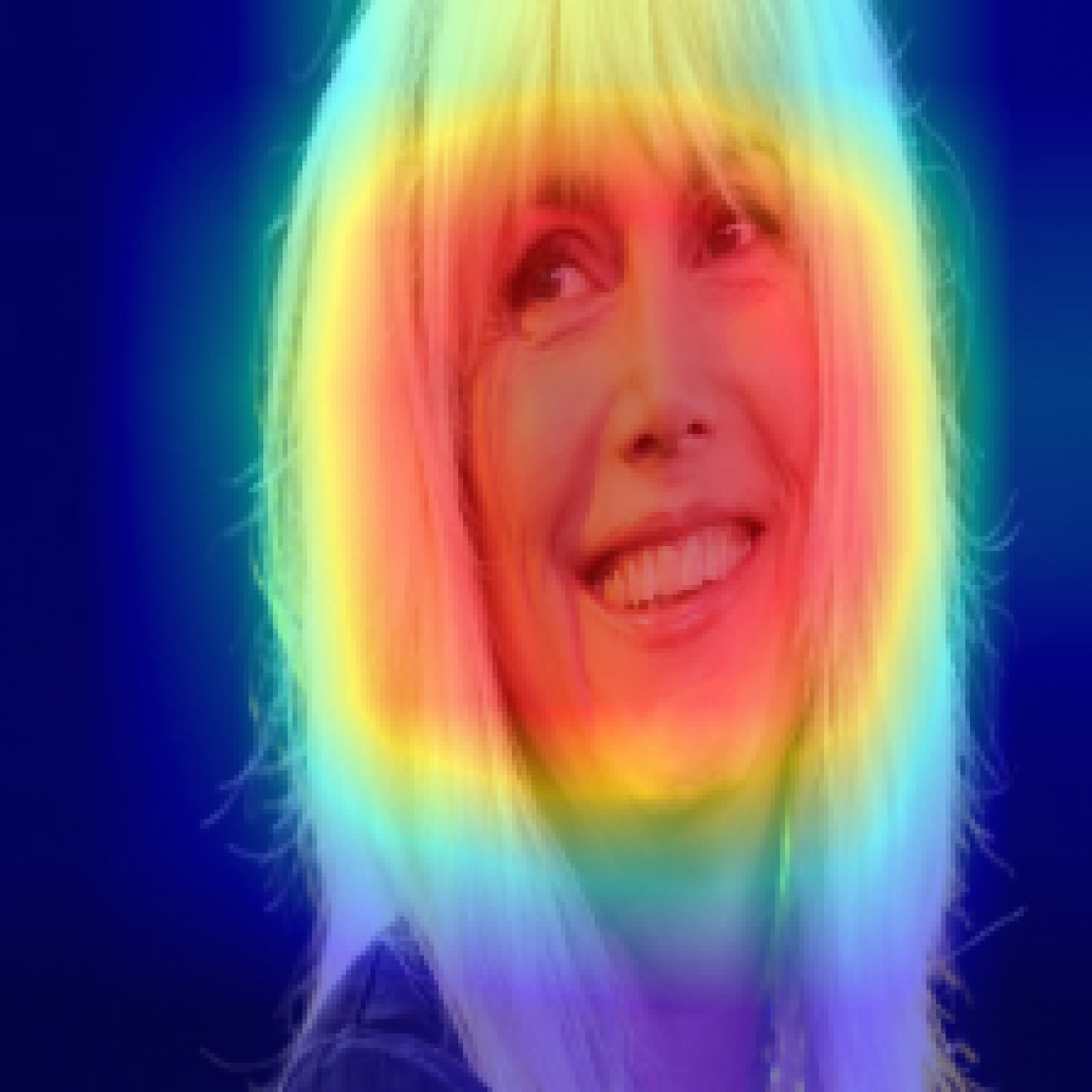}}
  \hfill
  \subfloat[]{\includegraphics[width=0.15\textwidth]{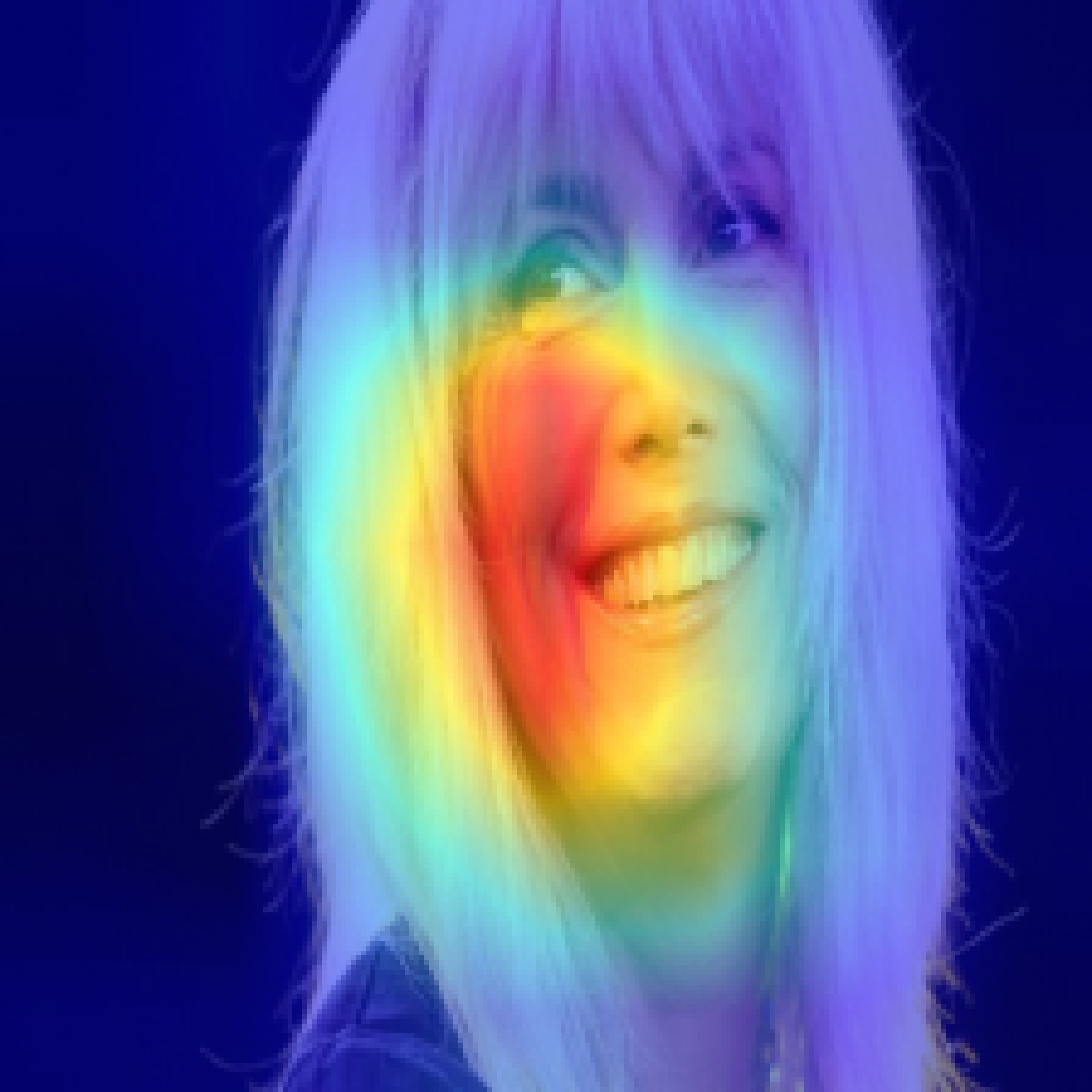}}
  \hfill
  \subfloat[]{\includegraphics[width=0.15\textwidth]{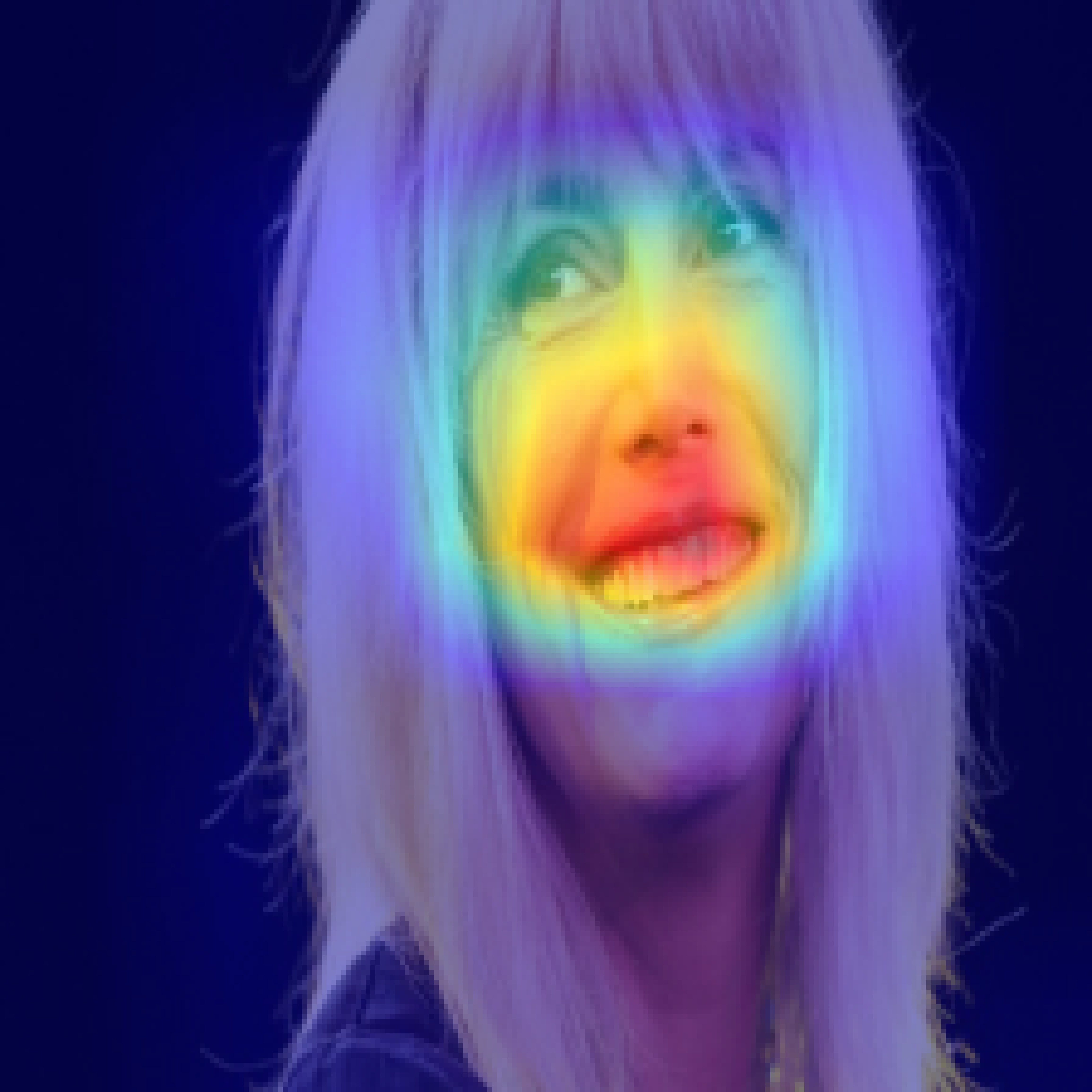}}
  
  \caption{Visualizing explanations using Grad-CAM for the proposed approach.`a/b' denotes the predicted/labeled information of each image, with `M/F' denoting male/female. (a) Age:1/1; (b) Gender:M/M; (c) Race:White/White; (d) Age:37/36; (e) Gender:M/M; (f) Race:Indian/Indian; (g) Age:42/60; (h) Gender:F/F; (i) Race:White/White.}
  \label{Grad}
\end{figure}

From the Figure~\ref{Grad}, we can draw the following conclusions.

The proposed approach exhibits a larger area of highlighted regions in the age estimation task, which may be attributed to a stronger correlation between age and features such as skin aging. The gender estimation task focuses on features such as jaw shape and facial hair, males often have facial hair. The race estimation task primarily focuses on the lower central region of the face ,such as the height of the nose bridge and skin color.

\subsection{Bias Study}

\begin{figure*}[h]
\centering
  \subfloat[]{\includegraphics[width=0.45\textwidth]{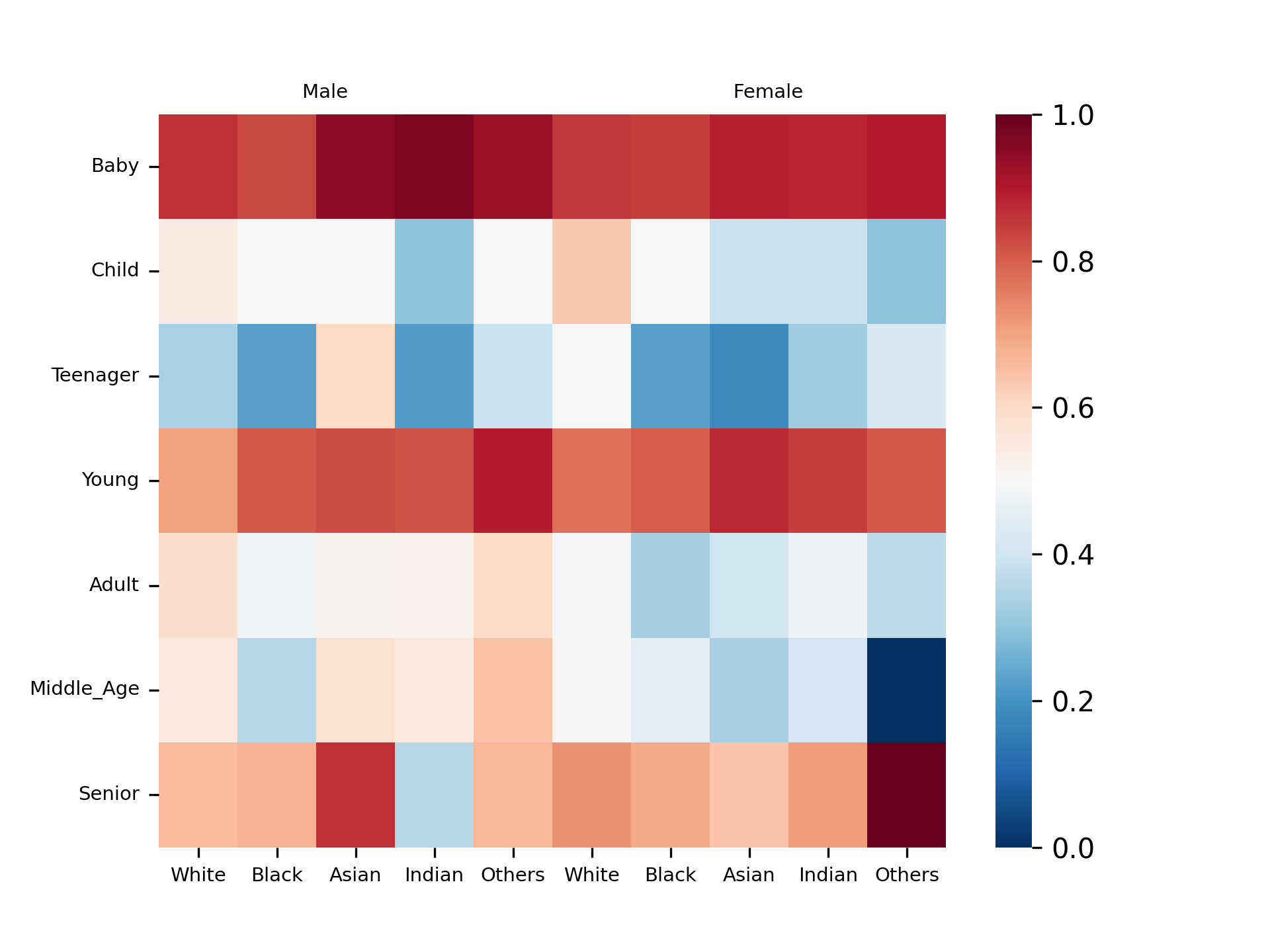}}
  \hfill
  \subfloat[]{\includegraphics[width=0.45\textwidth]{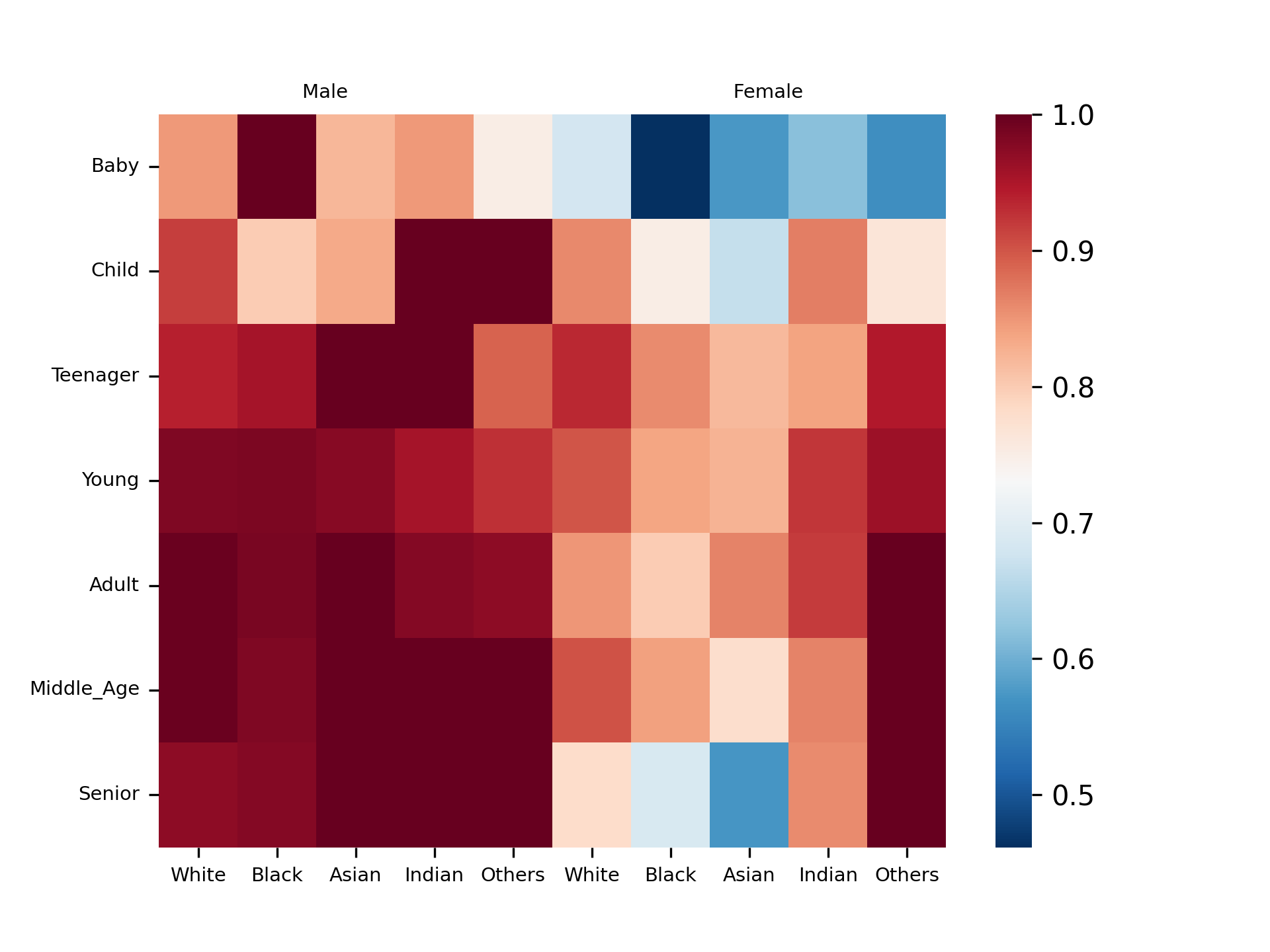}}
  \newline
  \subfloat[]{\includegraphics[width=0.45\textwidth]{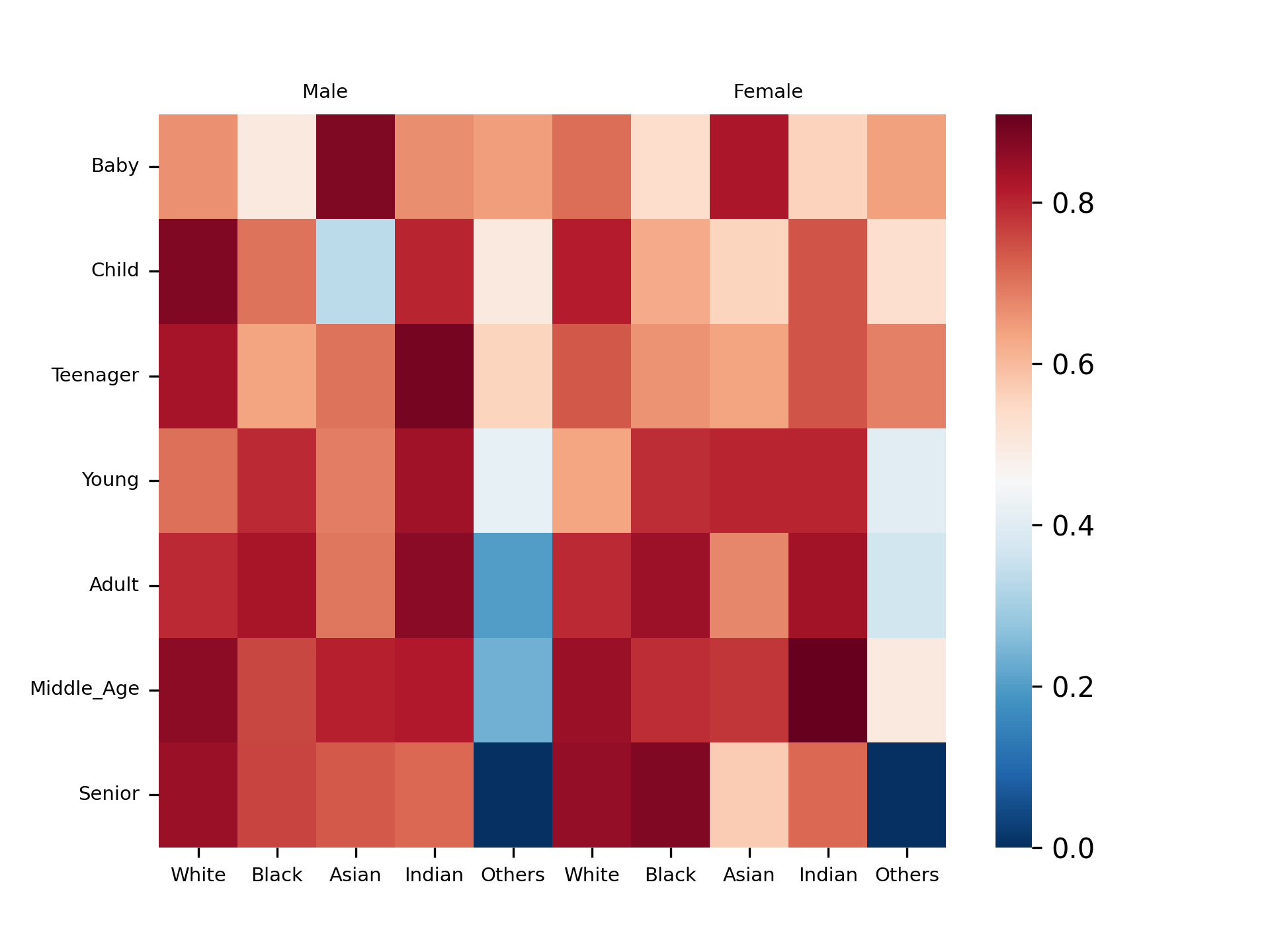}}
  \hfill
  \subfloat[]{\includegraphics[width=0.45\textwidth]{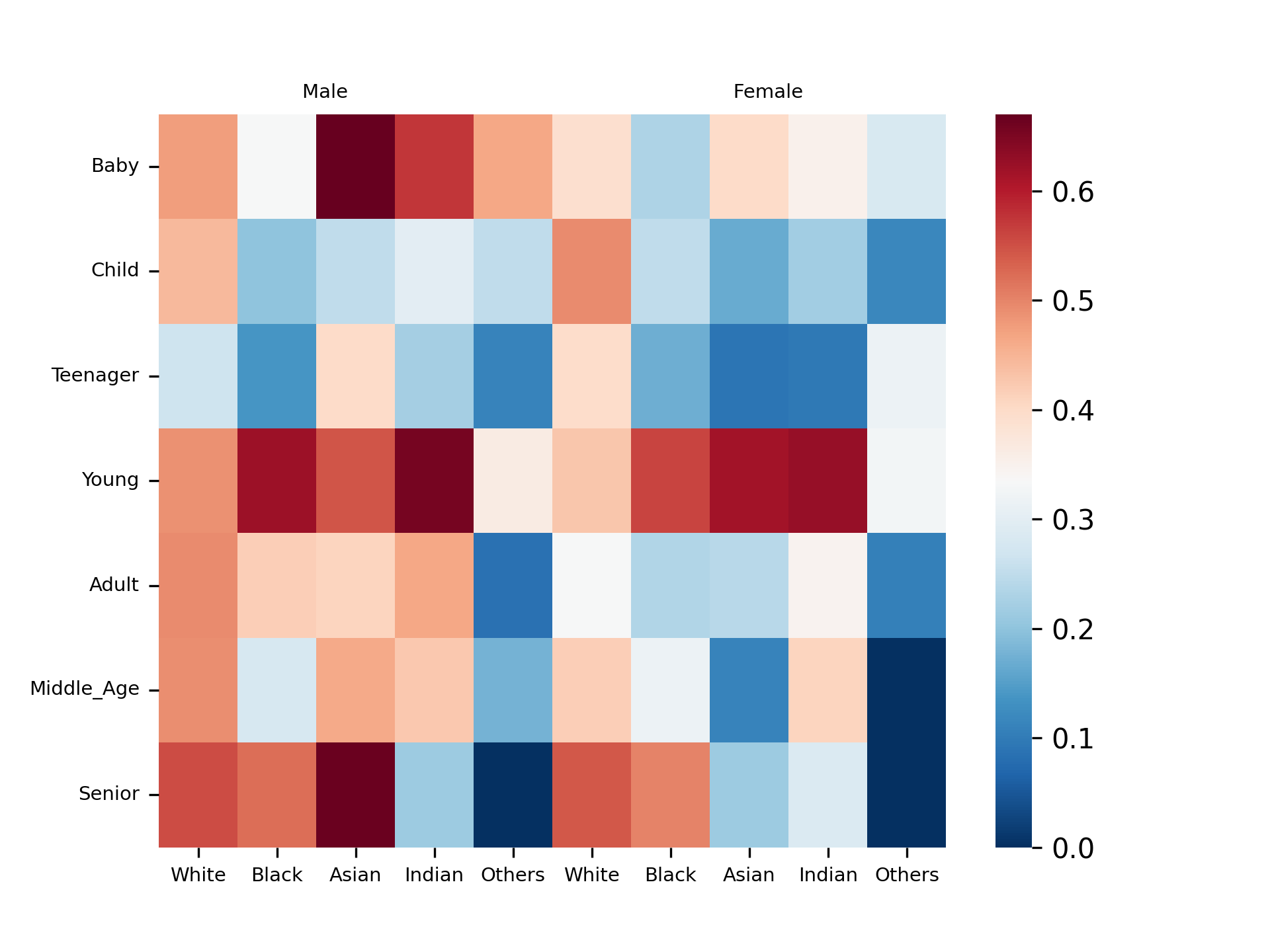}}
  
  \caption{Confusion matrix for face attribute estimation of the proposed approach on the UTKFace benchmark.}
  \label{confu_matrix}
\end{figure*}

The confusion matrix for face attributes along intersections of all attributes on the UTKFace benchmark is illustrated in Figure~\ref{confu_matrix}. It can be concluded that age, gender, and race estimation performance achieved by the proposed approach is rather encouraging.

However, due to the constraints of dataset imbalance and limitations of the approach itself, there are scenarios where the performance is relatively lower. The accuracy for the ``Teenager'' age group is lower in age estimation task, possibly due to highly overlapping features with adjacent age groups (``Child'', ``Young''). Gender estimation for females is more challenging, attributed to adornments and makeup. The lower accuracy in race estimation for Asian children may be attributed to less distinct skin color features.

\subsection{Engineering Practice}

\begin{figure}
\begin{center}
\includegraphics[width=2.5in]{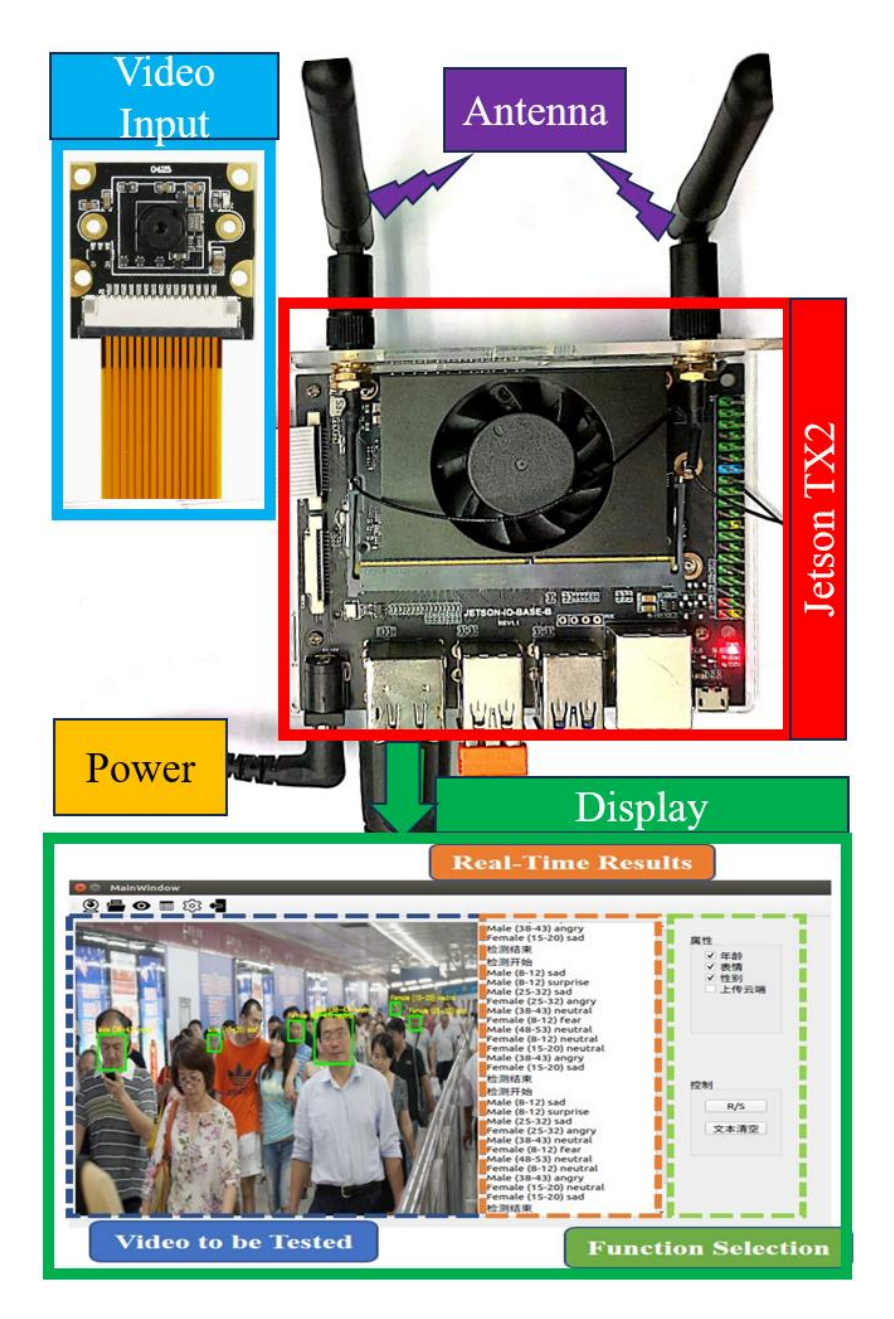}

\end{center}
\caption{The proposed approach was deployed and practically applied on Nvidia Jetson Tx2, and software was developed for interaction.}
\label{embed}

\end{figure}

As shown in  Figure \ref{embed}, we utilize a camera on an embedded gpu to capture video streams, and the proposed algorithm is hardware-accelerated based on TensorRT to achieve operation on low-end devices.  The test results are stored in a back-end database and the operator has access to this information through the provided back-end software.

The relevant software configurations and performance are provided in Table~\ref{T4-2}, where the lightweight neural network and TensorRT technology provide real-time guarantees.

\begin{table}[h]
\centering
\caption{Algorithm performance under different hardware and configurations}

\begin{tabular}{ccccccc}
\toprule[1.5pt]
Hardware & Framework  & Size & Time(s) & FPS & Error\\

\midrule[1pt]
Tesla K40c & PyTorch & 21M & 0.028  & 35.7 & -\\
Jetson TX2 & PyTorch  & 21M & 2.12  & 0.47 & -\\
Jetson TX2 & TensorRT & \textbf{16M} & \textbf{0.018}  & \textbf{55} & \textbf{1.138e-11} \\

\bottomrule
\end{tabular}
\label{T4-2}

\end{table}

\section{Conclusion}
In this paper, we propose a DMTL approach for facial attribute estimation. In order to account for both commonalities and differences among heterogeneous attributes, we implemented feature sharing and fine-tuning based on hard parameter sharing. The estimation error of ordinal attributes is reduced using a series of linear combinations of binary classifications problems.  The dynamic optimization of loss weights among multiple tasks is solved based on homoskedastic uncertainty.  The experimental results indicate that the proposed approach performs exceptionally well and is applicable to edge systems. Finally, we discuss interpretability and biases.

\section*{Acknowledgments}
This work was supported by the Tianjin Municipal Science and Technology Program (23KPXMRC00310).



 
\bibliographystyle{IEEEtran}
\bibliography{main}

\newpage

\vspace{11pt}

\begin{IEEEbiography}[{\includegraphics[width=1in,height=1.25in,clip,keepaspectratio]{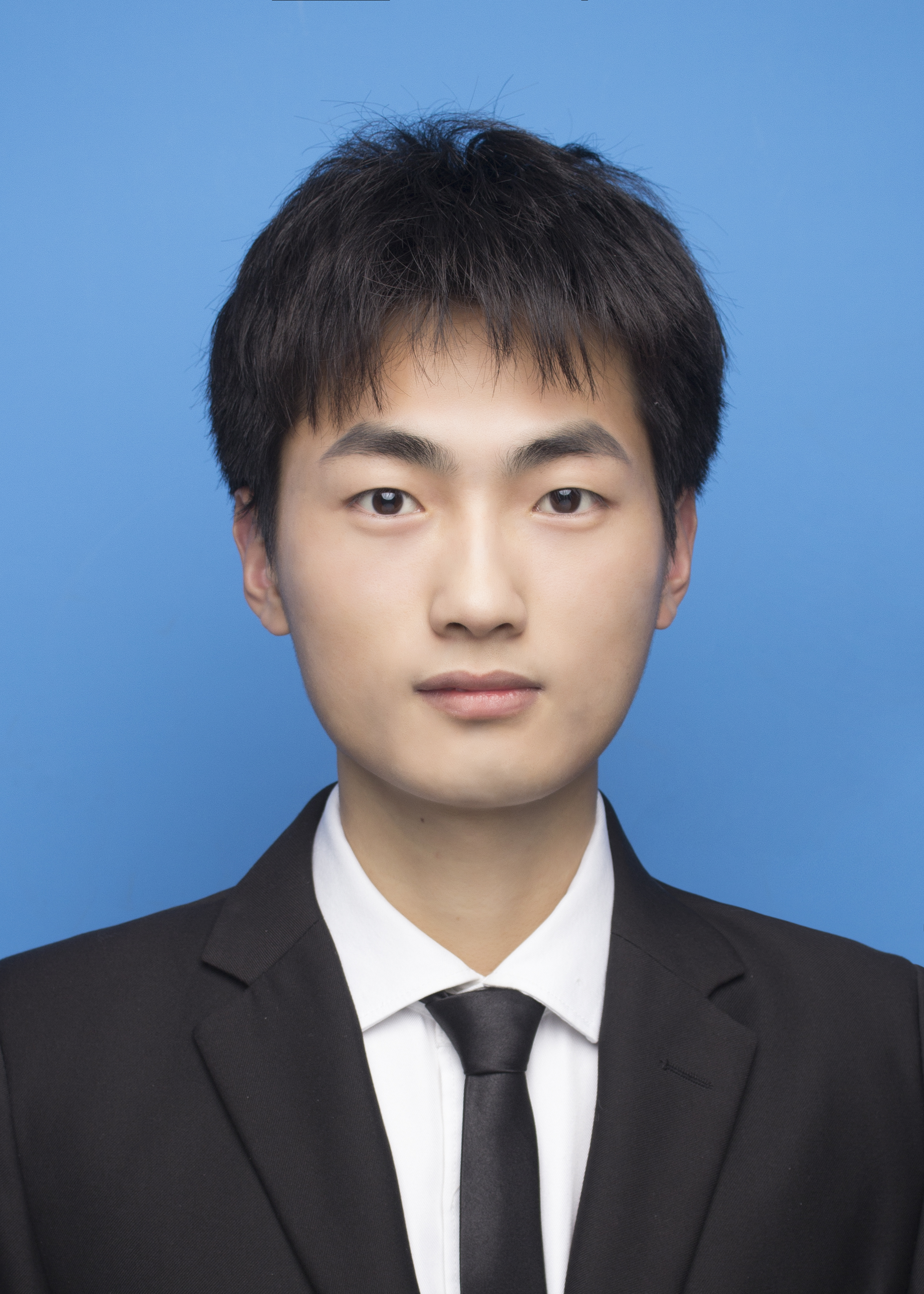}}]{Huaqing Yuan}
received the B.S. degree from the School of Information Engineering, Southwest University of Science and Technology, in 2021. He is currently pursuing the M.S. degree with the School of Electrical and Information Engineering, Tianjin University. His research interests include computer vision and multi-task learning.

\end{IEEEbiography}

\begin{IEEEbiography}[{\includegraphics[width=1in,height=1.25in,clip,keepaspectratio]{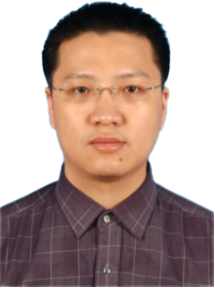}}]{Yi He}
received the B.S., M.S.degrees in control science and engineering and the Ph.D. degree in power electronics and power transmission from Tianjin University, Tianjin, China, in 2001, 2004, and 2007, respectively. He is currently an Associate Professor in School of Electrical and Information Engineering, Tianjin University. His research interests include machine learning and computer vision, deep learning ,industrial internet trustworthy system.

\end{IEEEbiography}

\begin{IEEEbiography}[{\includegraphics[width=1in,height=1.25in,clip,keepaspectratio]{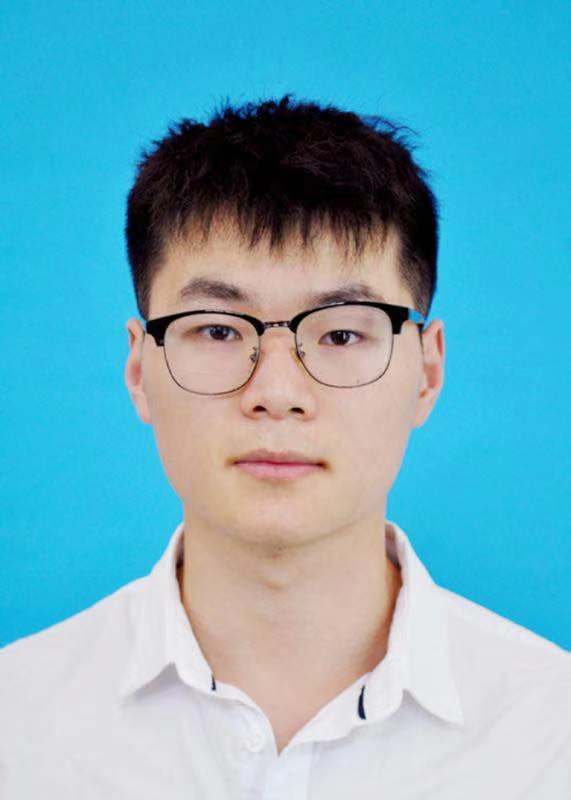}}]{Peng Du}
received his B.S. degree at Engineering College from Ocean University of China, Qingdao, China, 2022. He is currently working toward the M.S. degree at the School of Electrical and Information Engineering, Tianjin University, Tianjin. His current research interests are visual processing techniques such as eye tracking and pupil segmentation.

\end{IEEEbiography}

\begin{IEEEbiography}[{\includegraphics[width=1in,height=1.25in,clip,keepaspectratio]{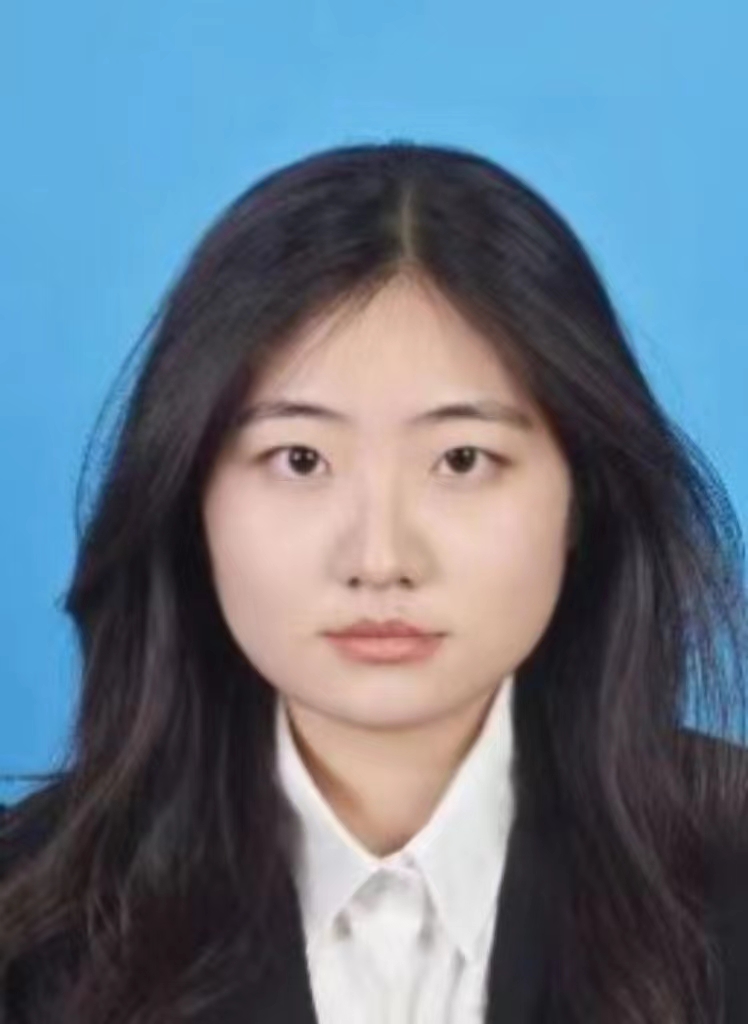}}]{Lu Song}
received the B.S. degree at Faculty of Information Technology from Beijing University of Technology, Beijing, China, 2023. She is currently working toward the M.S. degree at the School of Electrical and Information Engineering, Tianjin University, Tianjin. Her current research interests are deep learning.

\end{IEEEbiography}


\vfill

\end{document}